\documentclass{article}

\usepackage{iclr2027_conference,times}
\iclrfinaltrue

\usepackage{amsmath,amsfonts,bm}

\def\eqref#1{equation~\ref{#1}}

\def\1{\bm{1}}

\DeclareMathAlphabet{\mathsfit}{\encodingdefault}{\sfdefault}{m}{sl}
\SetMathAlphabet{\mathsfit}{bold}{\encodingdefault}{\sfdefault}{bx}{n}

\usepackage{wrapfig}
\usepackage[]{hyperref}
\usepackage{url}
\usepackage{booktabs}
\usepackage{amsmath}
\usepackage{caption}
\usepackage{bbm}
\usepackage{amssymb}
\usepackage{graphicx}
\usepackage{float}
\usepackage[table]{xcolor}
\usepackage{algorithm}
\usepackage{algorithmic}
\usepackage{lineno}
\usepackage{array}
\usepackage{multirow}
\usepackage{mhchem}
\usepackage{enumitem}
\usepackage[most]{tcolorbox}

\newtcolorbox{prompt}[1]{
    enhanced,
    breakable,
    colback=blue!3,
    colframe=black!70,
    boxrule=0.5pt,
    arc=2mm,
    left=10pt,
    right=10pt,
    boxsep=5pt,
    fonttitle=\bfseries,
    title=#1
}

\newtcolorbox{case}[1]{
    enhanced,
    breakable,
    colback=blue!3,
    colframe=blue!70,
    boxrule=0.5pt,
    arc=2mm,
    left=10pt,
    right=10pt,
    boxsep=5pt,
    fonttitle=\bfseries,
    title=#1
}

\newtcolorbox{sample}[1]{
    enhanced,
    breakable,
    colback=blue!3,
    colframe=teal!60,
    boxrule=0.5pt,
    arc=2mm,
    left=10pt,
    right=10pt,
    boxsep=5pt,
    fonttitle=\bfseries,
    title=#1
}

\def\iclrruler#1{}
\newcommand{\OURS}{ChemDeduce}
\title{\OURS{}: Benchmarking LLMs for Unknown Chemical Identification through Interactive Simulated Experiments}

\author{Hengyi Cai$^{1,2}$, Yuqi Tang$^{1,2}$, Zichang Li$^{1,2}$, Huajun Chen$^{1,2}$, Qiang Zhang$^{1,2}$, Keyan Ding$^{1,2}$\\
  $^1$Zhejiang University \;
  $^2$ZJU-Hangzhou Global Scientific and Technological Innovation Center\\
}

\begin{document}

\ificlrfinal\else
\linenumbers
\fi

\maketitle

\begin{abstract}
Large language models (LLMs) have shown strong performance in chemical knowledge and reasoning, yet it remains unclear whether they can solve problems whose answers must be discovered through experimentation. We investigate this question through Unknown Chemical Identification, where a model must actively design experiments, interpret observations, update hypotheses, and ultimately identify a hidden substance. We introduce \OURS{}, an interactive evaluation framework for assessing LLMs through simulated experiments. \OURS{} formulates chemical identification as a sequential decision process and provides a stateful simulated laboratory with explicit instrument and operational constraints, enabling scalable and reproducible evaluation of iterative experimental reasoning. We benchmark eight frontier LLMs on 200 chemicals, yielding $1{,}600$ model--target pairs and $4{,}800$ episodes over three independent repetitions, and evaluate identification accuracy, operational compliance, interaction efficiency, and trajectory-level reasoning quality. Despite generating seemingly plausible experimental plans, current models achieve an average identification accuracy ranging from 23.83\% to 67.33\%, revealing substantial limitations in experiment selection, evidence interpretation, and sequential hypothesis refinement.

\end{abstract}

\section{Introduction}

Large language models (LLMs) have demonstrated strong capabilities in chemical knowledge, molecular reasoning, synthesis planning, and chemistry-oriented tool use \citep{mirza2025framework,huang2026chemeval,m2024augmenting,yu2024chemtoolagent}. Yet many laboratory problems cannot be solved by recalling or reasoning over existing knowledge alone; their answers must instead be discovered through real-world experimentation. Unknown chemical identification provides a natural example. Given an unlabeled sample and a set of available reagents and instruments, an experimenter must select informative experiments, execute them under valid conditions, interpret the resulting observations, and progressively refine competing hypotheses based on accumulated evidence. This raises a fundamental question: \emph{can an LLM identify an unknown chemical by autonomously designing and conducting experiments?} Addressing this question requires an LLM to iteratively plan experiments, take actions, and interpret observations within a closed-loop process.

Existing chemistry benchmarks provide limited insight into this capability. Benchmarks such as ChemLLMBench, ChemBench, and ChemEval primarily evaluate static chemical knowledge, offline reasoning, or predefined tool use  \citep{guo2023can,mirza2025framework,huang2026chemeval}. In such settings, the information required for reasoning is largely available beforehand. Unknown chemical identification is fundamentally different: the model must actively acquire missing evidence through a sequence of experiments and use that evidence to progressively reduce uncertainty. Evaluating this process requires tracking the whole interactive experimental trajectory.

Such evaluation requires an environment that can support faithful and reproducible experimental interactions. The environment must respond to each intermediate operation with chemically plausible and state-consistent observations, so that later decisions depend on earlier actions. Real wet-lab experiments offer high physical fidelity but are costly and difficult to scale across many substances and models. Conversely, relying on an unconstrained LLM-only procedure to generate outcomes risks hallucinated phenomena and evaluator-dependent bias. A scalable benchmark therefore needs a constrained simulated environment that preserves the sequential structure of experimentation while controlling states, valid operations, and observable outcomes.

To this end, we introduce \textbf{\OURS{}}, an interactive evaluation framework for benchmarking LLMs on unknown chemical identification through simulated experiments. We formulate the task as a sequential decision process. In each episode, a target chemical is sampled from a pool of $200$ chemical substances and its identity is hidden from the model. Acting as an autonomous experimenter, the model proposes experimental operations in natural language, receives the resulting observations or instrument measurements, and iteratively gathers and interprets evidence before making a final identification under a limited interaction budget.

At the core of \OURS{} is \textbf{LabAgent}, a stateful laboratory simulator that integrates virtual instruments, reaction simlution, sample-state tracking, and operational constraints. LabAgent translates natural-language experimental instructions into executable actions and generates observations according to the current experimental state. It also leverages curated chemical knowledge bases and a domain-specific skill library to support reaction and instrument simulation. The stateful design enables reproducible multi-step experimentation while preserving causal consistency across the entire interaction trajectory.

Our contributions are threefold:
\begin{itemize}
\item {We introduce an interactive benchmark for unknown chemical identification.} Unlike existing static chemistry benchmarks, \OURS{} evaluates whether LLMs can actively design experiments, acquire informative evidence, and iteratively refine hypotheses, providing a controlled testbed for closed-loop experimental reasoning.
\item {We develop a stateful simulated laboratory for scalable and reproducible evaluation.}
The environment integrates virtual instruments, reaction simulation, sample-state tracking, and operational constraints, enabling natural-language experimental actions to produce consistent and state-dependent observations.
\item {We conduct a systematic evaluation of frontier LLMs as autonomous experimental agents.}
Across $1{,}600$ model--target pairs and $4{,}800$ episodes, our evaluation reveals substantial limitations in current models' ability to select informative experiments, interpret experimental evidence, and consistently update hypotheses over multiple rounds.
\end{itemize}

\section{Related Work}
\label{sec:related}

\paragraph{Static Evaluation in Chemistry.}
A growing body of work has evaluated the scientific and chemical capabilities
of LLMs through standardized benchmarks
\citep{hendrycks2020measuring,rein2023gpqa,lu2022learn,wang2023scibench}.
General-purpose benchmarks such as MMLU, GPQA, ScienceQA, and SciBench include
chemistry questions at undergraduate and graduate levels, while ChemLLMBench
\citep{guo2023can},
ChemBench
\citep{mirza2025framework},
ChemEval
\citep{huang2026chemeval},
ChemLLM
\citep{zhang2024chemllm},
and FGBench
\citep{liu2026fgbench}
more specifically assess chemical knowledge, textual and molecular reasoning,
safety awareness, and functional-group understanding.
Related work has also explored tool-augmented chemistry agents and LLM-based
scientific assistants for chemical reasoning, synthesis planning, and database
interaction
\citep{boiko2023autonomous,yu2024chemtoolagent,m2024augmenting}.
These benchmarks and systems provide standardized and reproducible comparisons
across models, but their evaluation protocols are predominantly static or
task-specific.
As a result, they capture important chemical competencies but provide limited
insight into whether a model can recognize information gaps, design informative
experiments, and acquire the evidence needed to resolve them.

\paragraph{Interactive Evaluation.}
Interactive benchmarks have increasingly been used to evaluate models in
settings where information must be acquired through sequential interaction,
tool use, and environment feedback
\citep{yao2022react,schick2023toolformer,liu2024agentbench,zhou2024webarena,wang2022scienceworld,jansen2024discoveryworld}.
AgentClinic
\citep{schmidgall2024agentclinic},
for example, evaluates agents in simulated clinical settings where they
interact with patients, request examinations, and progressively gather
evidence for diagnosis.
More general agent benchmarks, including ToolSandbox
\citep{lu2025toolsandbox},
AppWorld
\citep{trivedi2024appworld},
AgentBench
\citep{liu2024agentbench},
WebArena
\citep{zhou2024webarena},
and OSWorld
\citep{xie2024osworld},
evaluate multi-step decision making through interactions with constrained,
stateful environments.
ScienceWorld
\citep{wang2022scienceworld}
and DiscoveryWorld
\citep{jansen2024discoveryworld}
move closer to scientific inquiry by requiring agents to perform actions,
collect observations, and solve science-oriented tasks in virtual environments.
Self-driving laboratory and robotic chemistry studies further demonstrate the
potential of closed-loop experimentation for scientific discovery and chemical
optimization
\citep{burger2020mobile,macleod2020self,hase2018chimera}.
However, these benchmarks and systems are not designed to systematically
evaluate experimental reasoning for chemical identification.
To our knowledge, \OURS{} is the first interactive benchmark specifically
designed to evaluate whether LLMs can identify unknown chemicals by
sequentially designing simulated experiments, interpreting state-dependent
observations, and refining hypotheses over multiple rounds of interaction.

\section{Methods}
\label{sec:method}
\begin{figure}[htbp]
\centering
\includegraphics[width=\linewidth]{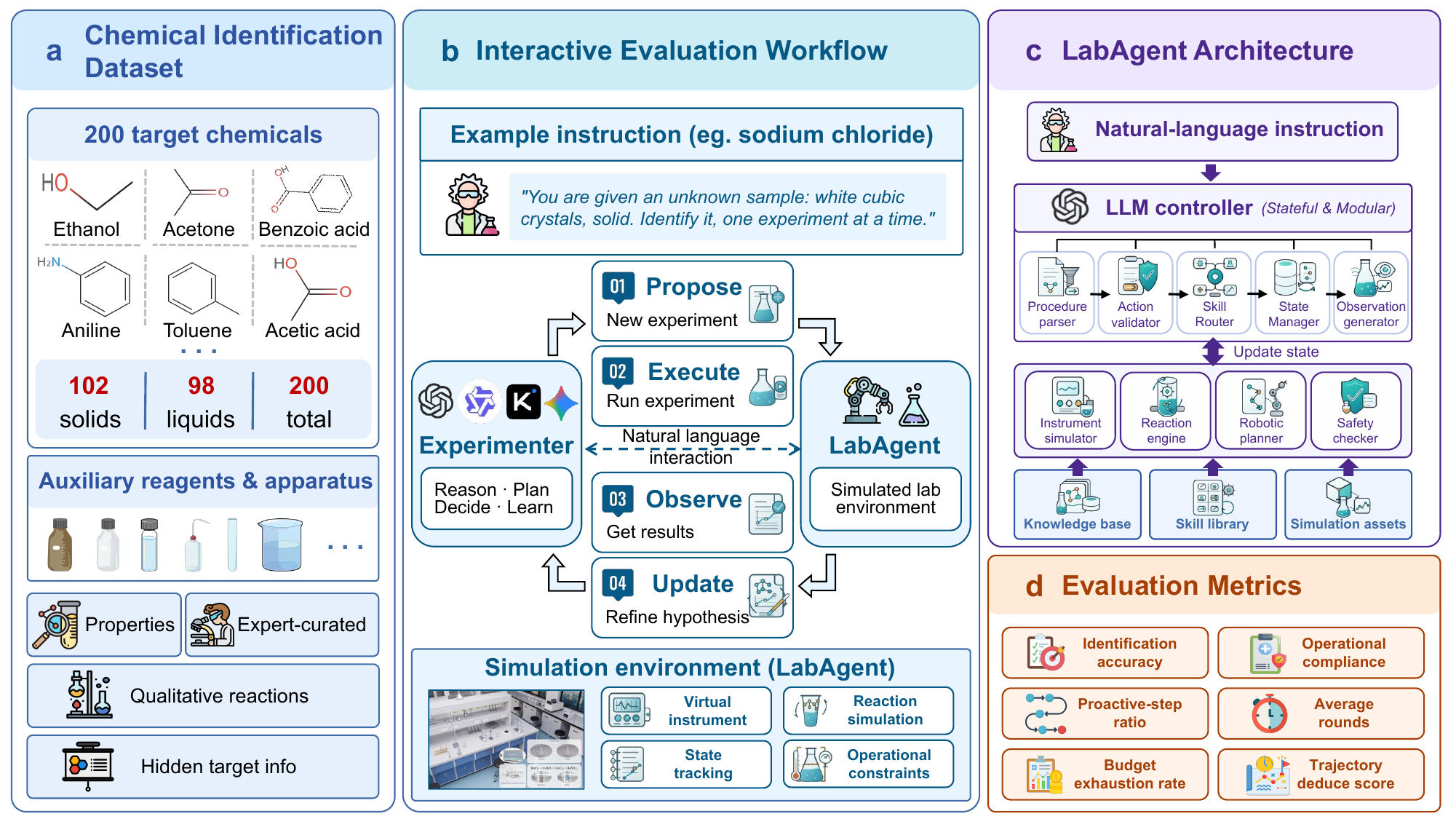}
\caption{Overview of \OURS{}.
(a) The Chemical Identification Dataset contains 200 target chemicals with curated experimental knowledge and virtual-laboratory resources.
(b) The Interactive Evaluation Workflow captures iterative experimentation, observation, and hypothesis refinement.
(c) LabAgent integrates an LLM Controller, Chemistry Skills, a Knowledge Base, a Robotic Manipulation Module, and Simulation Assets to execute experiments and generate state-consistent feedback.
(d) Evaluation Metrics cover identification accuracy, laboratory-operation validity, successful observation acquisition, interaction efficiency, budget exhaustion, and trajectory-level experimental reasoning quality.}
\label{fig:overview}
\end{figure}

In this section, we introduce \OURS{}, an interactive evaluation framework for unknown chemical identification, organized into four main parts: (1) a chemical identification dataset, (2) an interactive evaluation workflow, (3) the LabAgent architecture, and (4) evaluation metrics, as illustrated in Figure~\ref{fig:overview}.


\subsection{Chemical Identification Dataset}\label{sec:dataset}
\paragraph{Data Collection.}
To support realistic and controlled unknown chemical identification experiments, we collect candidate chemicals, laboratory reagents and associated experimental knowledge from \emph{Analytical Chemistry}~\citep{christian1977analytical}, \emph{The Systematic Identification of Organic Compounds}~\citep{shriner2003systematic}, \emph{Vogel's Textbook of Practical Organic Chemistry}~\citep{furniss2004practical}, and the \emph{CRC Handbook of Chemistry and Physics}~\citep{haynes2016crc}. 
The collected information mainly covers basic physicochemical properties, experimental characteristics, and representative qualitative reactions, together with the reagents and apparatus required for virtual experiments. These data provide the basis for LabAgent to update laboratory states and generate observable outcomes.

\paragraph{Dataset Generation.}
We normalize and structure the collected information by standardizing key fields and checking data completeness and consistency of experimental outcomes, and organize the resulting information into target-conditioned records for each chemical. We further retain only substances for which at least one viable identification route exists under the constraints of the simulated laboratory. We then invite three experts with chemistry backgrounds to independently review the samples according to three criteria: Chemical Correctness, Experimental Plausibility, and Record Consistency. Only chemicals approved by the majority vote are retained.

\paragraph{Final Dataset.}
Following the generation and verification process, we construct a dataset of 200 real chemicals, including 102 solids and 98 liquids, together with validated reagents, apparatus, and experimental records. The dataset covers diverse inorganic and organic chemicals, including several closely related substances that support identification tasks of varying difficulty. During evaluation, the candidate library and target-conditioned records remain hidden from the model under evaluation.

\subsection{Interactive Evaluation Workflow}\label{sec:workflow}
Inspired by the structure of real-world chemical identification workflows, \OURS{} models unknown chemical identification as a continuous interactive process that progressively acquires evidence through state-dependent experiments. This design moves beyond static chemistry evaluation by modeling the closed loop of experiments, observations, and hypothesis updates to assess iterative experimental reasoning in LLMs.


\paragraph{Agent Roles.}
Real-world chemical identification involves several participants with clearly delineated responsibilities. To reflect this separation, \OURS{} defines two core roles in the interactive environment:
\begin{equation}
\mathcal{A}=\{A_{\mathrm{Exp}},A_{\mathrm{Lab}}\}.
\label{eq:agent_roles}
\end{equation}
The Experimenter, an LLM denoted $A_{\mathrm{Exp}}$, represents the model under evaluation. It is responsible for all deliberative decisions and must identify the hidden target $m^*$ through iterative experimentation and reasoning over experimental observations. In contrast, LabAgent $A_{\mathrm{Lab}}$ functions as a responsive experimental environment that interprets, validates, and executes experimental requests, returning observable outcomes according to the current laboratory state. Together, these roles support the evaluation of iterative experimental reasoning.


\paragraph{Interaction Dynamics.}
Unknown chemical identification is modeled as a temporally ordered multi-turn experimental interaction:
\begin{equation}
H_t=(o_0,u_1,o_1,\ldots,u_t,o_t),
\qquad t=1,2,\ldots,T,\quad T\leq T_{\max}.
\label{eq:interaction_history}
\end{equation}
Here, $H_t$ denotes all information up to round $t$. Each episode is initialized with a hidden target $m^*$, and the Experimenter receives its physical state and macroscopic appearance as the initial observation $o_0$. At each subsequent round, the Experimenter issues a turn-level output $u_t$ based on the history $H_{t-1}$ and receives a new observation $o_t$ from LabAgent $A_{\mathrm{Lab}}$. This recursive process grounds subsequent decisions in accumulated experimental evidence, supporting iterative hypothesis refinement and unknown chemical identification.


\paragraph{Agent Actions.}
To characterize active experimental decision-making, the Experimenter selects an action from a discrete action space based on the interaction history:
\begin{equation}
\mathcal{A}_{\mathrm{Exp}}=\{\mathrm{Test},\mathrm{Identify}\}.
\label{eq:action_space}
\end{equation}
At round $t$, the Experimenter produces an action and its associated content conditioned on the interaction history:
\begin{equation}
(a_t^{\mathrm{Exp}},u_t)
=
A_{\mathrm{Exp}}(H_{t-1}),
\qquad
a_t^{\mathrm{Exp}}\in\mathcal{A}_{\mathrm{Exp}}.
\label{eq:agent_action}
\end{equation}
The action \emph{Test} yields $u_t=q_t$, corresponding to a natural-language experimental request for acquiring new observations. \emph{Identify} produces $u_t=\hat{m}$, representing the final chemical identity and terminating the episode. This abstraction effectively captures the core decision structure of chemical identification. 
The LabAgent $A_{\mathrm{Lab}}$ instead acts as the environment response mechanism, returning feedback according to the Experimenter's action $a_t^{\mathrm{Exp}}$ and the current laboratory state $s_t$.


\paragraph{LabAgent Response.}
Once the Experimenter issues a \emph{Test} request $q_t$, LabAgent $A_{\mathrm{Lab}}$ checks its executability under the current laboratory state $s_t$ and experimental rules $R$:
\begin{equation}
c_t=\mathrm{Check}(q_t,s_t,R)
\in\{\mathrm{Valid},\mathrm{Invalid}\},
\label{eq:validity_check}
\end{equation}
where $R$ specifies the available resources, safety constraints, and operational requirements. It then produces the observation $o_t$ and, for valid test requests, updates the laboratory state.:
\begin{equation}
o_t=
\begin{cases}
A_{\mathrm{Lab}}(q_t\mid s_t,m^*),
& \text{if } a_t^{\mathrm{Exp}}=\mathrm{Test}
\land u_t=q_t
\land c_t=\mathrm{Valid},\\
\mathrm{Reject},
& \text{if } a_t^{\mathrm{Exp}}=\mathrm{Test}
\land u_t=q_t
\land c_t=\mathrm{Invalid},\\
\mathrm{End},
& \text{if } a_t^{\mathrm{Exp}}=\mathrm{Identify}
\land u_t=\hat{m}.
\end{cases}
\label{eq:lab_response}
\end{equation}
A valid \emph{Test} updates the laboratory state and returns an observation, whereas an invalid request is rejected and leaves the state unchanged. Each response $o_t$ is constrained by the hidden target $m^*$ and current laboratory state $s_t$, ensuring state-consistent experimental feedback. This design grounds subsequent Experimenter reasoning in reliable experimental evidence. For valid test requests,  LabAgent also updates the laboratory state from $s_t$ to $s_{t+1}$ according to the executed operation; invalid requests return a rejection message and keep $s_{t+1}=s_t$.


\paragraph{History Update.}
After each interaction, the history is updated recursively as
\begin{equation}
H_t = H_{t-1} \oplus (u_t, o_t).
\label{eq:history_update}
\end{equation}
This operation appends the Experimenter's output $u_t$ and the corresponding LabAgent feedback $o_t$ to the interaction history. The updated $H_t$ is then used for subsequent decisions, enabling continuous hypothesis refinement from accumulated experimental results. The interaction terminates when $a_t^{\mathrm{Exp}}=\mathrm{Identify}$ or the interaction budget $T_{\max}$ is reached. The resulting trajectory is used for subsequent outcome evaluation and process analysis.


\subsection{LabAgent Architecture}\label{sec:lab_agent}
Unknown chemical identification requires state-dependent experimental interactions that provide reliable and state-consistent feedback. To this end, we construct LabAgent as the stateful experimental environment in \OURS{}. Given an experimental request $q_t$ and the current laboratory state $s_t$, LabAgent transforms the request into an executable experimental process and returns feedback $o_t$ consistent with the hidden target $m^*$ and the current state. To realize this process, LabAgent integrates five coordinated components: an LLM Controller, Chemistry Skills, a Knowledge Base, a Robotic Manipulation Module, and Simulation Assets.


\paragraph{LLM Controller.}
The LLM Controller interprets high-level experimental intent and coordinates LabAgent modules. In LabAgent, it parses the Experimenter's natural-language request into a structured experimental procedure specifying experimental objects, key parameters, and operation order, while organizing subsequent validation, tool invocation, and execution according to the current laboratory state. As the system's orchestration core, the LLM Controller coordinates modules under a shared laboratory state, ensuring consistency and coherence across multi-step experiments.


\paragraph{Chemistry Skills.}
Chemistry Skills encapsulate reusable chemical operations and domain-specific tools. During experiment execution, these skills invoke appropriate tools based on the structured procedure and current laboratory state, combining PubChem data with RDKit-based computation to determine state changes and observable outcomes. This grounds experimental feedback in explicit tool use and chemical computation, ensuring computability and reproducibility.


\paragraph{Knowledge Base.}
Knowledge Base supplements empirical knowledge that is not fully covered by general-purpose chemistry tools. It stores expert-validated substance properties and qualitative reaction records, which are retrieved when needed according to the hidden target, experimental conditions, and current laboratory state to support reliable outcome generation. To prevent information leakage, all target-specific records remain inaccessible to the Experimenter throughout evaluation.


\paragraph{Robotic Manipulation Module.}
The Robotic Manipulation Module translates high-level experimental procedures into concrete, executable laboratory operations. In LabAgent, it decomposes validated procedures into basic operations such as sampling, mixing, and measurement, while checking their execution preconditions against the available equipment and current laboratory state. 


\paragraph{Simulation Assets.}

Simulation Assets instantiate the virtual laboratory environment and its observable phenomena. Our simulation environment builds on LabUtopia~\citep{li2026labutopia}, which provides the underlying virtual-laboratory framework. They provide experimental scenes, devices, instruments, materials, and corresponding chemical effects, and combine the robotic action sequence with outcomes generated by Chemistry Skills to construct the experimental process and visual feedback. In this way, abstract laboratory state changes are translated into directly observable and recordable phenomena, providing the Experimenter with consistent experimental feedback. More implementation details and task-specific extensions are provided in Appendix~\ref{app:environment}.


The above components jointly form LabAgent's stateful experimental execution pipeline, transforming the Experimenter's natural-language request into a validated experimental procedure, chemically grounded outcome, executable operation, and final observable feedback. Moreover, LabAgent continuously updates the laboratory state and records key execution information, ensuring experimental consistency and supporting subsequent fine-grained analysis.


\subsection{Evaluation Metrics}\label{sec:metrics}

To comprehensively evaluate LLMs for unknown chemical identification, we design a multi-faceted evaluation protocol inspired by real laboratory identification workflows. Beyond final identification accuracy, the protocol further evaluates experimental execution, interaction efficiency, and trajectory-level reasoning quality, providing a comprehensive assessment of both identification outcomes and the underlying experimental process. Specifically, the evaluation metrics are described as follows.

\begin{enumerate}
    \item \textbf{Identification Accuracy:} The accuracy of the final identification, measuring whether the experimenter correctly identifies the hidden target. It provides a primary indicator of the model's identification performance.

    \item \textbf{Operational Compliance:} The rate at which experimental operations satisfy the required laboratory constraints, reflecting adherence to valid experimental procedures.

    \item \textbf{Proactive-Step Ratio:} The fraction of attempted operations that are successfully executed and return valid observations, capturing how effectively the model converts experimental requests into usable feedback.

    \item \textbf{Mean Interaction Rounds:} The number of interaction rounds until episode termination, indicating the efficiency of evidence gathering and decision completion.

    \item \textbf{Budget Exhaustion Rate:} The rate of episodes that reach $T_{\max}$ without a valid final declaration, reflecting the model's ability to complete identification within the interaction budget.

    \item \textbf{Trajectory Deduction Score:} The quality of the Experimenter's experimental reasoning trajectory, assessed by ``LLM-as-a-Judge'' across five dimensions: Candidate and Hypothesis Management (CHM), Diagnostic Experiment Design (DED), Observation Interpretation Fidelity (OIF), Evidential Sufficiency and Uniqueness (ESU), and Evidence-Gated Decision Efficiency (EGDE). Detailed definitions and scoring rubrics for each dimension are provided in Appendix~\ref{app:pds_details}.
\end{enumerate}

\section{Experiments}
\label{sec:experiments}

\subsection{Experimental Setup}
\label{sec:setup}

\paragraph{Models.}
We evaluate eight frontier models:
GPT-5.6-Sol \citep{openai2026gpt56},
Qwen3.8-Max~\citep{qwen2026qwen38},
Gemini-3.7-Flash \citep{gemini2026flash},
GLM-5.3 \citep{glm2026},
GLM-5.2 \citep{glm202652},
MiniMax-M3 \citep{lai2026minimax},
Grok-4.3 \citep{xai2026grok43}, and
Kimi-K3 \citep{team2026kimi}.
More details about these models are presented in Appendix~\ref{app:model_description}.

 \begin{table}[ht]
    \centering
    \small
    \caption{Unknown chemical identification results (mean $\pm$ standard error) on \OURS{}, including Acc. (identification accuracy), Op. Comp. (operational compliance), PSR (proactive-step ratio), Avg. Rounds (mean interaction rounds), BER (budget exhaustion rate), and TDS (trajectory deduction score).}
      \label{tab:main}
      \setlength{\tabcolsep}{4pt}
      \begin{tabular}{l>{\columncolor{gray!8}}cccccc}
      \toprule
      \textbf{Model} & \textbf{Acc.} $\uparrow$ & \textbf{Op. Comp.} $\uparrow$ &
      \textbf{PSR} $\uparrow$ & \textbf{Avg.\ rounds} $\downarrow$ &
      \textbf{BER} $\downarrow$ &
      \textbf{TDS} 
      $\uparrow$ \\
      \midrule
      GPT-5.6-Sol       & 67.33 \scriptsize{$\pm$ 1.77} & 88.29 \scriptsize{$\pm$ 0.42} & 81.71 \scriptsize{$\pm$ 0.64} & 6.97 \scriptsize{$\pm$ 0.05} & 16.50 \scriptsize{$\pm$ 1.80} &
      65.06 \scriptsize{$\pm$ 0.94}\\
      Qwen3.8-Max       & 60.67 \scriptsize{ $\pm$ 1.36} & 80.43 \scriptsize{ $\pm$ 0.34} & 76.14 \scriptsize{$\pm$ 0.49}& 6.88
      \scriptsize{$\pm$ 0.08}& 17.00 \scriptsize{$\pm$ 1.26} & 64.11 \scriptsize{$\pm$ 1.05} \\
        Gemini-3.7-Flash  & 55.00 \scriptsize{$\pm$ 0.50} & 71.88 \scriptsize{$\pm$ 1.14} &
        69.98 \scriptsize{$\pm$ 0.63} & 6.94 \scriptsize{$\pm$ 0.13} &
        12.33 \scriptsize{$\pm$ 1.45} & 55.26 \scriptsize{$\pm$ 0.32} \\
       GLM-5.3   & 48.83 \scriptsize{$\pm$ 0.93} & 75.45 \scriptsize{$\pm$ 0.70} & 71.13 \scriptsize{$\pm$ 0.74} & 8.16 \scriptsize{$\pm$ 0.06} & 38.00 \scriptsize{$\pm$ 1.15} &
       60.34 \scriptsize{$\pm$ 0.66}\\
      GLM-5.2  & 39.50 \scriptsize{$\pm$ 0.29} & 64.96 \scriptsize{$\pm$ 0.66} & 61.73 \scriptsize{$\pm$ 0.64} & 8.67 \scriptsize{$\pm$ 0.08} & 49.50 \scriptsize{$\pm$ 0.29} &
      54.95 \scriptsize{$\pm$ 0.69}\\
      MiniMax-M3    & 33.17 \scriptsize{$\pm$ 0.73} & 64.08 \scriptsize{$\pm$ 0.32} & 59.93 \scriptsize{$\pm$ 0.47} & 8.22 \scriptsize{$\pm$ 0.10} & 41.00 \scriptsize{$\pm$ 1.04} &
      46.15 \scriptsize{$\pm$ 0.18}\\
      Grok-4.3     & 29.17 \scriptsize{$\pm$ 1.09} & 60.22 \scriptsize{$\pm$ 0.16} & 56.54 \scriptsize{$\pm$ 0.28} & 8.33 \scriptsize{$\pm$ 0.03} & 37.83 \scriptsize{$\pm$ 1.02} &
      39.31 \scriptsize{$\pm$ 0.48}\\
      Kimi-K3    & 23.83 \scriptsize{$\pm$ 1.02} & 79.53 \scriptsize{$\pm$ 0.98} & 73.47 \scriptsize{$\pm$ 0.55} & 8.73 \scriptsize{$\pm$ 0.02} & 60.17 \scriptsize{$\pm$ 0.66} &
      44.17 \scriptsize{$\pm$ 0.55}\\
      \bottomrule
      \end{tabular}
    \end{table}
    
\paragraph{Evaluation protocol.}


To ensure a fair and reproducible evaluation, we adopt a unified experimental setting for all models. Specifically, each Experimenter model is evaluated on all $200$ target samples, yielding $1{,}600$ model--target pairs, with three independent runs to reduce sampling randomness, resulting in $4{,}800$ episodes in total. LabAgent uniformly uses DeepSeek-V4-Pro~\citep{xu2026deepseek} as the LLM controller. Model context and laboratory state are reset between episodes to ensure independence across evaluations. During Experimenter inference, the temperature is set to $0.3$, with a maximum output length of $8{,}192$ tokens. At the beginning of each episode, the Experimenter receives the public laboratory resources together with the target sample's initial physical state and macroscopic appearance, but not the target identity or target-conditioned records. Each episode ends upon final identification or after $T_{\max}=10$ interaction rounds. 
Unless otherwise specified, each reported mean is averaged over the three independent repetitions, and the standard error is computed across these repetitions.

\subsection{Main Results}\label{sec:main}
Table~\ref{tab:main} presents the unknown chemical identification results of eight LLMs on \OURS{}. The evaluation reveals a clear hierarchy in performance, with identification accuracy ranging from 23.83\% to 67.33\%. Notably, even frontier LLMs remain below 70\% accuracy under our framework, highlighting the challenge of sequential chemical identification. Interestingly, Kimi-K3 achieves 79.53\% operational compliance and a 73.47\% proactive-step ratio, exceeding Gemini-3.7-Flash and GLM-5.2 on these execution-oriented metrics, yet attaining the lowest identification accuracy. This indicates that stronger experimental execution does not necessarily translate into better identification without effective interpretation and integration of experimental evidence.

Moreover, interaction rounds are not consistently associated with higher identification accuracy. Qwen3.8-Max was the most efficient, requiring only 6.88 rounds on average, yet achieved only moderate identification accuracy. The top-performing model, GPT-5.6-Sol, demonstrated a more effective and robust strategy, whereas Kimi-K3 required the most interactions while attaining the lowest accuracy. This suggests that interaction efficiency depends on how effectively each experiment narrows the candidate space rather than simply increasing interactions. Meanwhile, budget exhaustion is more pronounced among weaker models, reaching 60.17\% in Kimi-K3 and 49.50\% in GLM-5.2, substantially exceeding that of stronger models, 
suggesting that weaker models are more likely to fail to reach a valid final declaration within the limited interaction budget.

\begin{wrapfigure}{r}{0.6\linewidth}
\vspace{-0.8em}
\centering
\includegraphics[width=1.0\linewidth]{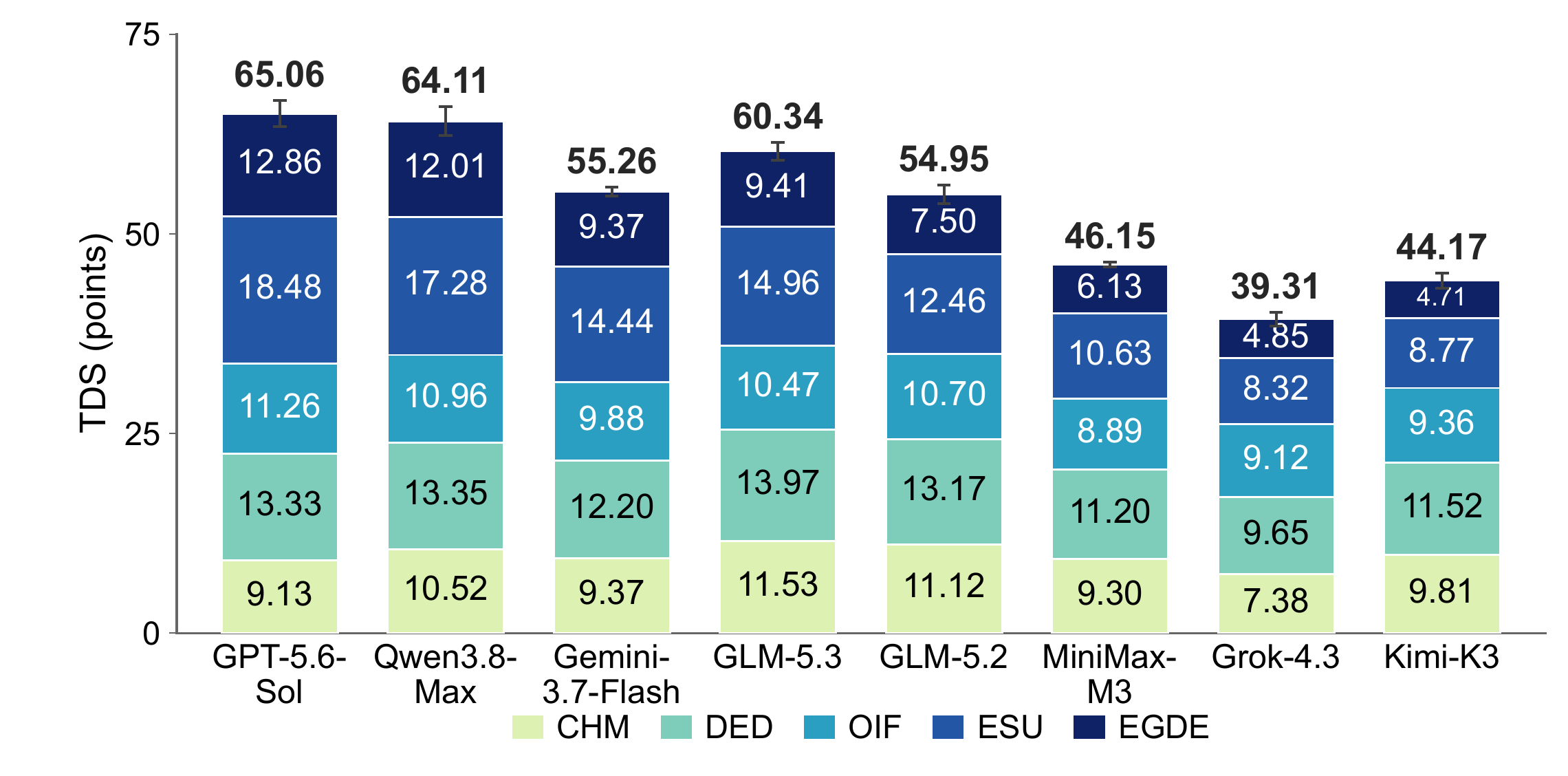}
\caption{Trajectory Deduction Score (TDS, out of 100) across five dimensions: CHM (15), DED (20), OIF (15), ESU (30), and EGDE (20).}
\label{fig:tds_by_model}
\vspace{-3em}
\end{wrapfigure}


Figure~\ref{fig:tds_by_model} further analyzes the experimental reasoning process across five TDS dimensions. 
A critical finding is that evidence integration and final decision-making remain prominent weaknesses across all evaluated LLMs. In terms of Evidential Sufficiency and Uniqueness (ESU) and Evidence-Gated Decision Efficiency (EGDE), model scores generally fall in the low to moderate range, far below their respective maximum scores. Even the strongest model, GPT-5.6-Sol, achieves only 18.48 out of 30 on ESU and 12.86 out of 20 on EGDE, indicating that current models still struggle to integrate multi-round observations into sufficiently discriminative evidence and translate that evidence into reliable final decisions. Notably, these two dimensions also exhibit the largest variation across models, with gaps of 10.16 and 8.11 points between the highest and lowest scores, respectively. This suggests that evidential sufficiency and the quality of final decisions are key factors that differentiate models in experimental reasoning. In contrast, most models show relatively stable performance on Candidate and Hypothesis Management (CHM), Diagnostic Experiment Design (DED), and Observation Interpretation Fidelity (OIF), with comparatively small variation across models, showing that current LLMs already possess relatively stable basic experimental reasoning capabilities.

\subsection{Failure Analysis}\label{sec:error}

\begin{figure}[ht]
\centering
\vspace{-0.35cm}
\setlength{\abovecaptionskip}{0cm}
\includegraphics[width=\linewidth]{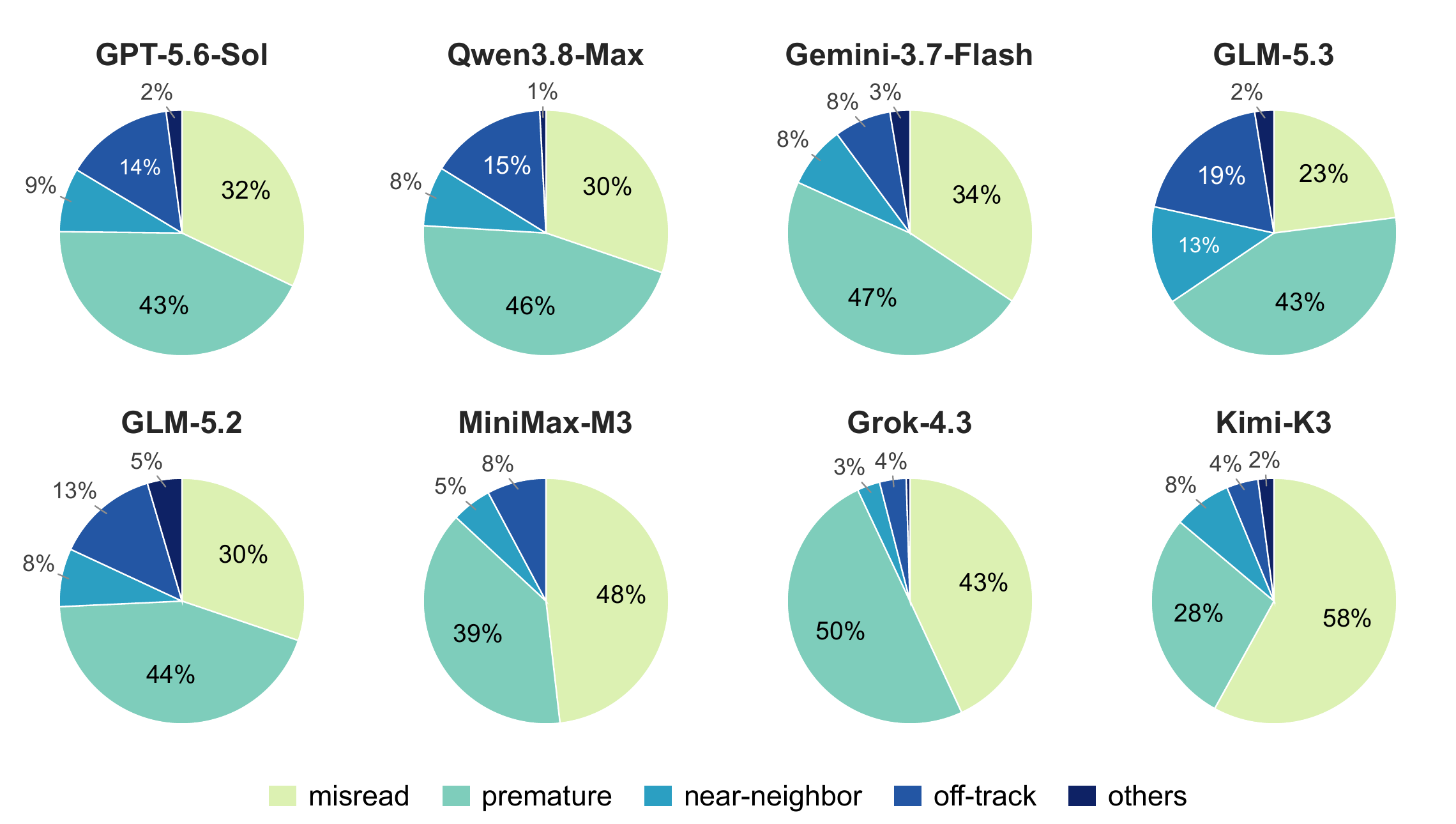}
\caption{Error type distribution among incorrect identifications across eight Experimenter models. Each pie shows the proportions of five categories: \emph{misread} (contradicted evidence), \emph{premature} (premature conclusion), \emph{near-neighbor} (close-relative error), \emph{off-track} (off-target search), and \emph{others}.}
  \label{fig:cause_pies}
\end{figure}

To characterize fine-grained error differences across models, we categorize incorrect declarations into five types: \emph{misread}, \emph{premature}, \emph{near-neighbor}, \emph{off-track}, and \emph{others}, reflecting failures in evidence handling, decision timing, and discrimination among similar candidates. As shown in Figure~\ref{fig:cause_pies}, \emph{misread} and \emph{premature} are the most prevalent error types across models, with average proportions of 37.3\% and 42.5\%, respectively, substantially higher than the 7.8\% and 10.6\% observed for \emph{near-neighbor} and \emph{off-track}. 
Notably, \emph{premature} errors are not driven by explicit pressure from the interaction budget, as the Experimenter is not informed of the exact round limit. These findings indicate that the primary bottleneck of current models lies in accurately interpreting and integrating available evidence and making reliable decisions only when the evidence is sufficient.

A fine-grained analysis across Experimenter models reveals distinct error patterns. Stronger models such as GPT-5.6-Sol and Qwen3.8-Max are dominated by \emph{premature} errors, whereas weaker models exhibit more \emph{misread} errors. In particular, \emph{misread} accounts for 58\% and 48\% of the errors made by Kimi-K3 and MiniMax-M3, respectively, highlighting difficulties in interpreting observations and resolving contradictory evidence. Interestingly, the GLM models (i.e., GLM-5.3 and GLM-5.2) exhibit the fewest \emph{misread} errors in absolute percentage, yet show a more pronounced tendency to end without a final declaration, indicating that fewer evidence-interpretation errors do not necessarily translate into stronger overall identification performance.

\subsection{Sensitivity of LabAgent Controller}\label{sec:LabAgent_Controller}
To further analyze the sensitivity of \OURS{} to the LabAgent controller, we keep all other evaluation settings unchanged and evaluate three Experimenter models under three LabAgent controller configurations: DeepSeek-V4-Pro, StepFun-3.7-Flash, and Qwen3.8-Flash. As shown in Figure~\ref{fig:ablation} (a), the relative ranking of the three models remains unchanged across all controllers, while the accuracy variation of each model is within 5.0 percentage points. These results suggest that \OURS{} is robust across different LabAgent controllers and provides consistent comparative conclusions across Experimenter models.


\subsection{Budget Sensitivity}\label{sec:budget}
\begin{figure}[ht]
\centering
\setlength{\abovecaptionskip}{0cm}
\includegraphics[width=\linewidth]{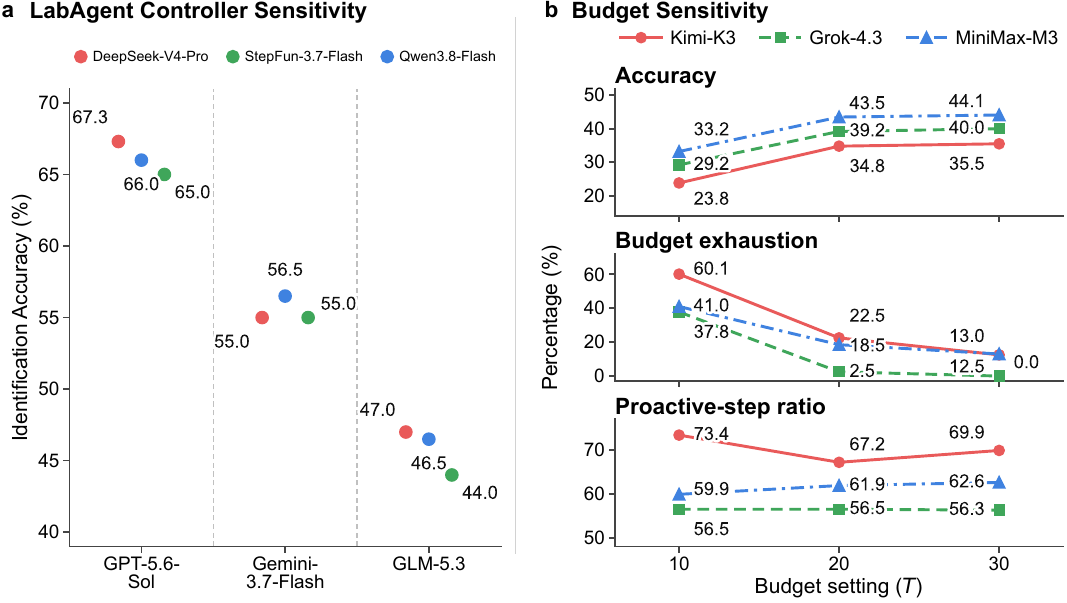}
\caption{Sensitivity analyses. (a) Identification accuracy of Experimenter models across three LabAgent controllers (DeepSeek-V4-Pro, Qwen3.8-Flash, and StepFun-3.7-Flash). (b) Identification performance of three Experimenter models under interaction budgets of $T_{\max}=10$, $20$, and $30$.}
\label{fig:ablation}
\end{figure}

To further examine the effect of interaction budget on \OURS{}, we evaluate three Experimenter models, Kimi-K3, Grok-4.3, and MiniMax-M3, under interaction budgets of $T_{\max}=10$, $20$, and $30$. These models show high budget exhaustion and low identification accuracy under the default $T_{\max}=10$ setting.
As shown in Figure~\ref{fig:ablation} (b), all three models exhibit a consistent trend. Increasing the budget from $T_{\max}=10$ to $T_{\max}=20$ improves identification accuracy and reduces budget exhaustion, whereas further increasing it to $T_{\max}=30$ yields little additional gain. This pattern suggests that increasing the interaction budget improves task completion
when the budget is restrictive, whereas beyond this point performance is limited more
by experiment selection and evidence integration. This supports the use of
$T_{\max}=10$ as our default resource-constrained setting: the budget is restrictive
enough to expose differences in trajectory quality, yet not so restrictive that success
is primarily determined by the interaction cap. The sensitivity analysis is included to
show how the evaluated budget-limited models respond when the constraint is relaxed.


\section{Conclusion}\label{sec:conclusion}
In this work, we propose an interactive simulated-experiment framework for evaluating LLMs on unknown chemical identification. \OURS{} moves beyond static chemistry benchmarks by formulating identification as a sequential experimental process in which models must actively design experiments, interpret observations, and update hypotheses from accumulated evidence. With 200 chemicals and a stateful simulated laboratory, \OURS{} enables controlled and reproducible evaluation of iterative experimental reasoning. Our experiments reveal that current frontier LLMs can generate plausible experimental plans but still struggle with informative experiment selection, evidence interpretation, and multi-round hypothesis refinement, highlighting a substantial gap between static chemical reasoning and closed-loop experimental reasoning.


\subsection*{AI Use Statement}

The language models evaluated in this work are the objects of study and were used within the research system under the experimental protocol described in the paper; this is separate from the use of generative AI tools for manuscript preparation. In preparing the manuscript, generative AI tools were used only for drafting assistance, language polishing, and formatting, and all AI-assisted text was reviewed and revised by the authors. These manuscript-preparation tools were not used to create, recompute, or modify the reported measurements, which were computed from recorded evaluation traces generated under the stated protocol. The authors take full responsibility for the correctness, integrity, and presentation of the final work.

\subsection*{Ethics Statement}

This research aims to improve the reliability and safety of language models in scientific reasoning and laboratory decision-making. All experiments in this work are conducted in a text-mediated simulated laboratory environment and do not involve human participants, personally identifiable information, or physical chemical experiments. While the benchmark is designed to reflect realistic laboratory operations and may help inform future real-world experimental workflows, it does not replace professional training, expert supervision, or institutional safety review. Any deployment in a physical laboratory should be conducted only under qualified human oversight and appropriate safety controls. We therefore foresee no significant ethical issues arising from this study.

\subsection*{Reproducibility Statement}

We are committed to supporting the reproducibility of our research. We will release our source code containing the simulated laboratory, the $200$-sample library, prompts, configuration files, and evaluation scripts. The paper provides details on the evaluated models, experimental protocol, prompts, and evaluation metrics. These resources will enable the community to inspect the benchmark protocol, reproduce the simulator-side evaluation pipeline, and replicate the reported analyses to the extent permitted by access to the same or comparable model APIs.

\bibliography{iclr2027_conference}
\bibliographystyle{iclr2027_conference}

\clearpage
\appendix
\section*{Appendix}
\setcounter{table}{0}
\setcounter{figure}{0}
\setcounter{section}{0}
\setcounter{equation}{0}
\renewcommand{\thetable}{A\arabic{table}}
\renewcommand{\thefigure}{A\arabic{figure}}
\renewcommand{\theequation}{A\arabic{equation}}

\section{Limitations}
\label{app:limitations}


While \OURS{} provides a simulated environment for evaluating laboratory-oriented scientific reasoning, several limitations remain. First, \OURS{} abstracts laboratory identification into text-based interaction, improving reproducibility but excluding multimodal perception from images, videos, instrument displays, and subtle visual cues. Second, observations are represented through qualitative phenomena and partial instrument readings whose precision is constrained by the simulator settings, which reduces over-interpretation but does not fully capture real-world variability such as contamination, reagent aging, instrument drift, calibration error, and manual handling. Future work could incorporate multimodal observations, finer-grained measurement models, and more realistic physical noise to further bridge the gap between simulated and real-world laboratories.

\section{Comparison with Existing Benchmarks}
\label{app:related_comparison}

Table~\ref{tab:related_comparison} summarizes the relationship between
\OURS{} and representative benchmarks for chemical reasoning, scientific
question answering, and interactive agent evaluation. Existing chemistry
benchmarks primarily evaluate static knowledge, molecular reasoning, or
predefined tool-use tasks. General interactive benchmarks evaluate sequential
decision-making but are not designed for unknown chemical identification.
\OURS{} differs by requiring a model to actively acquire evidence through
stateful simulated experiments and to identify a hidden chemical under
operational constraints.

\begin{table}[ht]
\centering
\small
\caption{Comparison between \OURS{} and representative chemistry, scientific,
and interactive-agent benchmarks. ``Simulated Env.'' indicates whether the benchmark
provides a simulated environment for interactive task execution. ``Stateful''
indicates whether the environment maintains persistent task state across turns.
``Experiment Design'' indicates whether the model must choose informative
experiments rather than answer static questions.}
\vspace{0.3em}
\label{tab:related_comparison}
\resizebox{\textwidth}{!}{%
\begin{tabular}{lccccccc}
\toprule
Benchmark & Domain & Simulated Env. & Static QA & Interactive & Stateful & Experiment Design & Chemical ID \\
\midrule
MMLU \citep{hendrycks2020measuring}
& General science & -- & \checkmark & -- & -- & -- & -- \\
GPQA \citep{rein2023gpqa}
& Graduate science & -- & \checkmark & -- & -- & -- & -- \\
ChemLLMBench \citep{guo2023can}
& Chemistry & -- & \checkmark & -- & -- & -- & -- \\
ChemBench \citep{mirza2025framework}
& Chemistry & -- & \checkmark & -- & -- & -- & -- \\
ChemEval \citep{huang2026chemeval}
& Chemistry & Limited & \checkmark & Limited & Limited & Limited & -- \\
ChemLLM \citep{zhang2024chemllm}
& Chemistry & -- & \checkmark & -- & -- & -- & -- \\
FGBench \citep{liu2026fgbench}
& Functional groups & -- & \checkmark & -- & -- & -- & -- \\
AgentBench \citep{liu2024agentbench}
& General agents & \checkmark & -- & \checkmark & \checkmark & -- & -- \\
WebArena \citep{zhou2024webarena}
& Web agents & \checkmark & -- & \checkmark & \checkmark & -- & -- \\
ToolSandbox \citep{lu2025toolsandbox}
& Tool use & \checkmark & -- & \checkmark & \checkmark & -- & -- \\
AppWorld \citep{trivedi2024appworld}
& App agents & \checkmark & -- & \checkmark & \checkmark & -- & -- \\
ScienceWorld \citep{wang2022scienceworld}
& Science tasks & \checkmark & -- & \checkmark & \checkmark & Partial & -- \\
DiscoveryWorld \citep{jansen2024discoveryworld}
& Scientific discovery & \checkmark & -- & \checkmark & \checkmark & Partial & -- \\
AgentClinic \citep{schmidgall2024agentclinic}
& Clinical diagnosis & \checkmark & -- & \checkmark & \checkmark & Diagnostic tests & -- \\
\midrule
\textbf{\OURS{}} & \textbf{Chemical identification}
& \textbf{\checkmark} & \textbf{--} & \textbf{\checkmark} & \textbf{\checkmark} & \textbf{\checkmark} & \textbf{\checkmark} \\
\bottomrule
\end{tabular}%
}
\end{table}

The comparison highlights three distinguishing properties of \OURS{}.
First, the answer is hidden and must be inferred from newly acquired
experimental evidence rather than recalled from a question statement. Second,
the environment is stateful: an operation changes sample, vessel, apparatus,
and observation states, which affect subsequent interactions. Third, evaluation
covers not only final-answer accuracy but also operational compliance, interaction
efficiency, budget exhaustion, and process-level evidence use.

\section{Dataset Details}
\label{app:dataset}

\subsection{Dataset Construction Pipeline}
\label{app:dataset_pipeline}

The \OURS{} dataset is constructed to support controlled unknown chemical
identification in a simulated laboratory. The construction pipeline consists of
four stages.

\paragraph{Source collection.}
We collect candidate substances and experimental knowledge from standard
analytical and practical chemistry references, including \emph{Analytical
Chemistry}, \emph{The Systematic Identification of Organic Compounds},
\emph{Vogel's Textbook of Practical Organic Chemistry}, and the \emph{CRC
Handbook of Chemistry and Physics}. The collected information includes physical
state, appearance, basic physicochemical properties, qualitative reactions,
available reagents, apparatus requirements, operating conditions, and observable
experimental outcomes.

\paragraph{Normalization and structuring.}
Raw records are converted into a unified structured schema. Chemical names,
aliases, physical states, class labels, reagent names, apparatus names, and
observation descriptions are normalized to reduce ambiguity. When multiple
sources describe the same test, we reconcile differences by prioritizing
standard qualitative-analysis conditions and expert-verified observations.

\paragraph{Expert review.}
Three reviewers with chemistry backgrounds independently evaluate candidate
records according to chemical correctness, experimental plausibility, and record
consistency. A chemical is retained only if it receives majority approval.
Records with ambiguous identities, unsupported observations, conflicting test
outcomes, or insufficient distinguishing information are revised or removed.

\paragraph{Simulator integration.}
The verified records are converted into target-conditioned knowledge profiles
used by LabAgent. These profiles specify hidden target identity, public initial
appearance, allowable observations, relevant qualitative reactions, instrument
responses, safety constraints, and waste-handling requirements. During evaluation, the complete target-conditioned profiles remain hidden from the
Experimenter.

\subsection{Dataset Statistics}
\label{app:dataset_stats}

The final benchmark contains $200$ real chemical substances, including $102$
solids and $98$ liquids. The library covers inorganic substances, elemental
samples, and common organic functional-group families.
Table~\ref{tab:class} reports the mean
identification accuracy of each model by chemical class. These results provide
a complementary view of the dataset by showing that different categories induce
substantially different levels of difficulty. For example, elemental substances
and inorganic salts are generally easier to identify, whereas halogenated
compounds, amines/amides, and several organic functional-group classes are more
challenging. This suggests that \OURS{} contains not only diverse chemical
categories, but also heterogeneous identification difficulty across categories.

  \begin{table}[H]
    \centering
    \small
    \caption{Identification accuracy ($\%$) by chemical class at $T_{\max}=10$,
    averaged over the three repeats. $n$ is the number of targets in the class
    and \emph{mean} averages the eight models. Solid and Liquid split the same 200
    targets by the sample's physical state.
    For each row, the best model score is shown in bold and the second-best
    score is underlined; ties are all marked.}
    \vspace{0.3em}
    \label{tab:class}
    \resizebox{\textwidth}{!}{%
    \begin{tabular}{lcccccccccc}
    \toprule
    Class & $n$ & GPT-5.6 & Qwen3.8 & Gemini & GLM-5.3 & GLM-5.2 & MiniMax & Grok-4.3 & Kimi-K3 & mean \\
    \midrule

    Halogenated
    & 11
    & \textbf{57.6}
    & \underline{48.5}
    & 42.4
    & 21.2
    & 24.2
    & 12.1
    & 9.1
    & 6.1
    & 27.7 \\

    Amine / amide
    & 22
    & \textbf{45.5}
    & \underline{39.4}
    & \underline{39.4}
    & 30.3
    & 22.7
    & 10.6
    & 9.1
    & 7.6
    & 25.6 \\

    Phenol
    & 10
    & \textbf{63.3}
    & 40.0
    & 36.7
    & \underline{46.7}
    & 30.0
    & 16.7
    & 30.0
    & 16.7
    & 35.0 \\

    Ether
    & 8
    & \textbf{58.3}
    & \underline{50.0}
    & 37.5
    & 37.5
    & 37.5
    & 33.3
    & 20.8
    & 16.7
    & 36.5 \\

    Ester
    & 13
    & \textbf{69.2}
    & \textbf{69.2}
    & \underline{43.6}
    & 20.5
    & 28.2
    & \underline{43.6}
    & 23.1
    & 10.3
    & 38.5 \\

    Carboxylic acid
    & 22
    & \textbf{60.6}
    & \underline{50.0}
    & 47.0
    & 47.0
    & 37.9
    & 21.2
    & 24.2
    & 19.7
    & 38.4 \\

    N,S-heterocycle
    & 10
    & \textbf{66.7}
    & \textbf{66.7}
    & 43.3
    & \underline{50.0}
    & 40.0
    & 36.7
    & 23.3
    & 10.0
    & 42.1 \\

    Alcohol
    & 22
    & \textbf{78.8}
    & \underline{71.2}
    & 63.6
    & 50.0
    & 33.3
    & 39.4
    & 27.3
    & 16.7
    & 47.5 \\

    Ketone
    & 11
    & \textbf{78.8}
    & 63.6
    & \underline{69.7}
    & 39.4
    & 15.2
    & 30.3
    & 6.1
    & 15.2
    & 39.8 \\

    Aldehyde
    & 10
    & \textbf{63.3}
    & \textbf{63.3}
    & 53.3
    & \underline{60.0}
    & 36.7
    & 40.0
    & 26.7
    & 23.3
    & 45.8 \\

    Carbohydrate
    & 8
    & \textbf{79.2}
    & 70.8
    & \underline{75.0}
    & 62.5
    & 50.0
    & 33.3
    & 45.8
    & 33.3
    & 56.3 \\

    Hydrocarbon
    & 12
    & \textbf{91.7}
    & \underline{69.4}
    & 63.9
    & 61.1
    & 58.3
    & 47.2
    & 41.7
    & 38.9
    & 59.0 \\

    Inorganic salt
    & 36
    & 67.6
    & \textbf{71.3}
    & 67.6
    & \underline{70.4}
    & 60.2
    & 44.4
    & 49.1
    & 47.2
    & 59.7 \\

    Elemental
    & 5
    & \underline{86.7}
    & 80.0
    & \textbf{93.3}
    & 80.0
    & 80.0
    & 80.0
    & \underline{86.7}
    & 66.7
    & 81.7 \\

    \midrule

    Solid
    & 102
    & \textbf{62.7}
    & \underline{56.2}
    & 54.6
    & 51.0
    & 42.8
    & 30.4
    & 35.3
    & 29.1
    & 45.3 \\

    Liquid
    & 98
    & \textbf{72.1}
    & \underline{65.3}
    & 55.4
    & 46.6
    & 36.1
    & 36.1
    & 22.8
    & 18.0
    & 44.0 \\

    \midrule

    All targets
    & 200
    & \textbf{67.3}
    & \underline{60.7}
    & 55.0
    & 48.8
    & 39.5
    & 33.2
    & 29.2
    & 23.8
    & 44.7 \\

    \bottomrule
    \end{tabular}%
    }
  \end{table}

\subsection{Dataset Record Examples}
\label{app:dataset_record_examples}

The \OURS{} dataset contains two types of structured records:
target-conditioned chemical records and laboratory resource records.
Target-conditioned records describe hidden substances, including their verified
names, aliases, physical states, macroscopic appearances, physicochemical
properties, qualitative reactions, and supported measurements. Laboratory
resource records describe available instruments, apparatus, reagents,
accessories, storage locations, usage constraints, and safety information.
The following examples illustrate the storage format used in the dataset.

\begin{sample}{Solid Sample Profile: Sodium Chloride}

\textbf{Basic Identity}
\begin{itemize}[left=0pt]
  \item \textbf{Name}: Sodium chloride
  \item \textbf{Aliases}: Table salt; NaCl; common salt
  \item \textbf{State}: Solid
\end{itemize}

\textbf{Physical Properties}
\begin{itemize}[left=0pt]
  \item \textbf{Melting Point}: $801^\circ\mathrm{C}$
  \item \textbf{Boiling Point}: Not available
  \item \textbf{Specific Rotation}: Not applicable
  \item \textbf{Refractive Index at $20^\circ\mathrm{C}$}: Not available
  \item \textbf{Conductivity at $25^\circ\mathrm{C}$}: $85.0\,\mathrm{mS/cm}$
  \item \textbf{IR Characteristic Peaks}: Not available
\end{itemize}

\textbf{Macroscopic Appearance}
\begin{itemize}[left=0pt]
  \item White cubic crystals
\end{itemize}

\textbf{Curated Qualitative Reactions}
\begin{itemize}[left=0pt]
  \item \textbf{Reagent / Operation}: Platinum wire flame test
  \begin{itemize}
    \item \textbf{Phenomenon}: The flame appears yellow.
  \end{itemize}

  \item \textbf{Reagent / Operation}: Silver nitrate solution
  \begin{itemize}
    \item \textbf{Phenomenon}: Adding silver nitrate solution to the aqueous
    solution produces a white precipitate. The precipitate does not dissolve
    after adding dilute nitric acid.
  \end{itemize}
\end{itemize}

\end{sample}

\begin{sample}{Liquid Sample Profile: Methanol}

\textbf{Basic Identity}
\begin{itemize}[left=0pt]
  \item \textbf{Name}: Methanol
  \item \textbf{Aliases}: Methanol; wood alcohol; industrial alcohol
  \item \textbf{State}: Liquid
\end{itemize}

\textbf{Physical Properties}
\begin{itemize}[left=0pt]
  \item \textbf{Melting Point}: $-98^\circ\mathrm{C}$
\item \textbf{Boiling Point}: $64.7^\circ\mathrm{C}$
\item \textbf{Refractive Index at $20^\circ\mathrm{C}$}: $1.3288$
\item \textbf{Conductivity at $25^\circ\mathrm{C}$}: Not available
  \item \textbf{IR Characteristic Peaks}: Not available
\end{itemize}

\textbf{Macroscopic Appearance}
\begin{itemize}[left=0pt]
  \item Colorless liquid with an ethanol-like odor
\end{itemize}

\textbf{Curated Qualitative Reactions}
\begin{itemize}[left=0pt]
  \item \textbf{Reagent / Operation}: Metallic sodium
  \begin{itemize}
    \item \textbf{Phenomenon}: Adding metallic sodium produces colorless gas
    bubbles.
  \end{itemize}

  \item \textbf{Reagent / Operation}: Lucas reagent
  \begin{itemize}
    \item \textbf{Phenomenon}: No turbidity is observed. As a primary alcohol,
    methanol does not react even upon heating.
  \end{itemize}

  \item \textbf{Reagent / Operation}: Iodoform test
  \begin{itemize}
    \item \textbf{Phenomenon}: Negative result. The sample does not contain the
    required structural motif for a positive iodoform reaction.
  \end{itemize}
\end{itemize}

\end{sample}

\begin{sample}{Instrument Profile: Refractometer}

\textbf{Basic Information}
\begin{itemize}[left=0pt]
  \item \textbf{Name}: Refractometer
  \item \textbf{Quantity}: 1
  \item \textbf{Storage Location}: Center of the instrument bench
\end{itemize}

\textbf{Required Accessories}
\begin{itemize}[left=0pt]
  \item Lens tissue
  \item Rubber dropper
\end{itemize}

\textbf{Sample Preparation}
\begin{itemize}[left=0pt]
  \item Use a rubber dropper to place $2$--$3$ drops of the liquid sample onto
  the prism surface of the refractometer.
  \item Close the prism cover.
  \item Adjust the reflector toward the light source until a clear bright--dark
  boundary appears in the field of view.
\end{itemize}

\textbf{Usage Constraints}
\begin{itemize}[left=0pt]
  \item The refractometer is applicable only to liquid samples.
  \item The prism surface must not contact hard objects or strongly corrosive
  liquids.
  \item After use, the prism surface must be gently cleaned with lens tissue
  moistened with ethanol.
\end{itemize}

\textbf{Notes}
\begin{itemize}[left=0pt]
  \item The test temperature is fixed at $25^\circ\mathrm{C}$.
  \item The measurement range is $n_D=1.3000$--$1.7000$.
\end{itemize}

\end{sample}

\begin{sample}{General Apparatus Profile: Beaker($250\,\mathrm{mL}$)}

\textbf{Basic Information}
\begin{itemize}[left=0pt]
  \item \textbf{Name}: Beaker, $250\,\mathrm{mL}$
  \item \textbf{Quantity}: 8
  \item \textbf{Storage Location}: Middle shelf of the apparatus cabinet
  \item \textbf{Material}: Borosilicate glass
\end{itemize}

\textbf{Role in the Virtual Laboratory}
\begin{itemize}[left=0pt]
  \item The beaker is treated as a general-purpose container for operations such
  as dissolving, mixing, temporary liquid transfer, and observation of
  macroscopic phenomena.
  \item Its availability is checked by LabAgent when an experimental procedure
  requires a generic vessel.
\end{itemize}

\end{sample}

\begin{sample}{Reagent Profile: Dilute Hydrochloric Acid}

\textbf{Basic Information}
\begin{itemize}[left=0pt]
  \item \textbf{Name}: Dilute hydrochloric acid
  \item \textbf{Aliases}: Dilute HCl; HCl(aq); hydrochloric acid
  \item \textbf{State}: Liquid
  \item \textbf{Quantity}: Sufficient
  \item \textbf{Concentration}: Approximately $2\,\mathrm{mol/L}$
\end{itemize}

\textbf{Container and Storage}
\begin{itemize}[left=0pt]
  \item \textbf{Container}: Transparent glass reagent bottle, $500\,\mathrm{mL}$,
  with a ground-glass stopper
  \item \textbf{Storage Location}: Acidic reagent area of the reagent rack
\end{itemize}

\textbf{Hazards}
\begin{itemize}[left=0pt]
  \item Corrosive liquid.
  \item Avoid contact with skin and eyes.
  \item Reacts with metals and releases hydrogen gas.
\end{itemize}

\textbf{Role in the Virtual Laboratory}
\begin{itemize}[left=0pt]
  \item This reagent can be selected by the Experimenter for supported
  acid-addition or acid-confirmation tests.
  \item LabAgent checks whether its use is compatible with the sample state,
  requested operation, apparatus, and safety constraints before execution.
\end{itemize}

\end{sample}

\subsection{Information Isolation}
\label{app:dataset_isolation}

The dataset contains both public and hidden information. Public information
includes the available laboratory resources, their usage constraints, and the
initial physical state and appearance of the unknown sample. Hidden information
includes the target identity, verified aliases, target-conditioned reaction
records, diagnostic cues, and answer-resolution records. During evaluation, the
Experimenter can only access public laboratory information and observations
returned by LabAgent, while LabAgent can access the hidden target-conditioned
records required for state-consistent simulation.

\section{Simulated Laboratory Environment}
\label{app:environment}

All experiments in \OURS{} are conducted in a simulated chemistry laboratory
built in NVIDIA Isaac Sim. The environment builds on the LabUtopia simulation
platform~\citep{li2026labutopia} and recreates a real laboratory bench with the
glassware and instruments that qualitative identification procedures require, so
that each requested experiment can be executed and recorded in a controlled
setting.

The simulation infrastructure, the robotic arm, and the layered interface for
defining and executing tasks all come from LabUtopia. Its chemistry assets,
however, are limited to a few common items such as beakers and measuring
cylinders, and its action vocabulary consists of generic manipulations such as
shaking and pouring, which do not cover the qualitative procedures we evaluate.
On top of the platform we add the chemistry layer: 30 chemical actions in the
action catalog that the robot can execute, and 70 pieces of equipment and 31
experiment scenes in the asset library, all assembled by generation scripts from
shared assets rather than modeled one by one.

The Cartesian controller provided by the platform (RMPflow) is not sufficient on
its own, because operations such as pushing a test tube back into its rack or
lowering a dropper into a bottle neck have to land within about a millimeter. We
therefore add an inverse-kinematics engine that solves each target pose and
validates the solution before the arm moves. Chemical phenomena are likewise
computed rather than replayed: color, liquid level, bubbles, precipitates and
layered phases are updated for every frame from the experiment parameters and the
current configuration of the arm. The same scene can thus produce different
outcomes under different parameters, while a fixed parameter set reproduces the
same video.

The asset library includes the equipment required by our bench protocols:
laboratory benches with shelves of reagent bottles, sinks, gas and power outlets;
glassware such as test tubes and racks, beakers, conical flasks, funnels,
droppers, glass rods, spatulas, and weighing bottles; instruments including a
balance, a refractometer, a conductivity meter, pH paper, and thermometers;
heating devices such as an alcohol lamp, a Bunsen burner, and a hot plate; and the reagents used in
the qualitative tests. The chemical substances are drawn from the sample library
described in Section~\ref{sec:dataset}. Because the apparatus is kept as a
reusable asset library, experiments that require common equipment are assembled
from shared assets rather than implemented as separate scenes.

\begin{figure}[ht]
    \centering
    \vspace{-0.5em}
    \includegraphics[width=\linewidth]{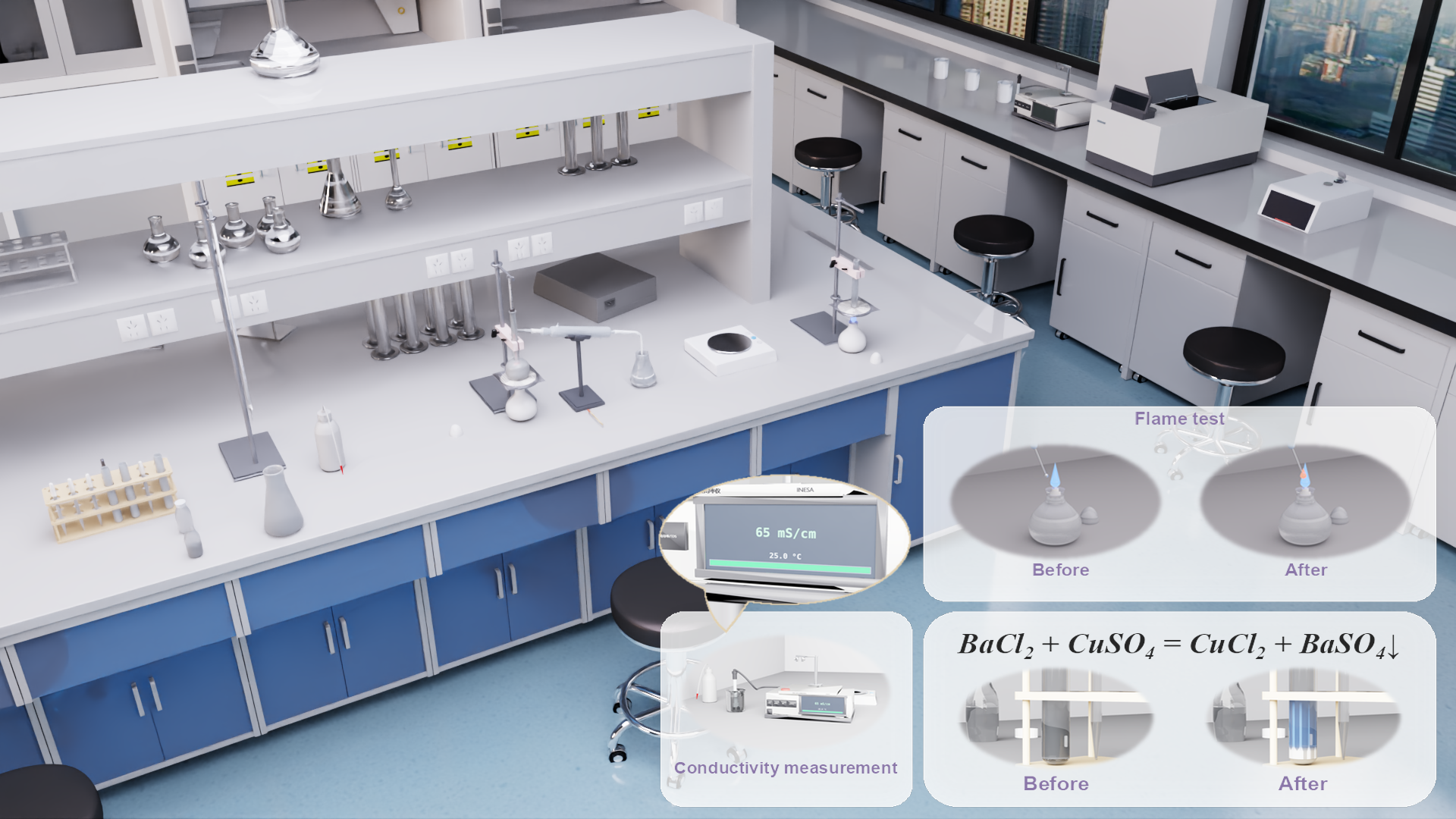}
    \caption{The simulated laboratory environment of \OURS{}. The overview camera
    covers the bench, its glassware and its instruments; the three call-outs show
    phenomena that the environment can produce and record in close-up, namely a
    conductivity reading, the flame of a flame test before and after the sample is
    introduced, and a precipitation reaction with an illustrative equation shown only for visualization.}
    \label{fig:app_env_overview}
\end{figure}

The environment supports five categories of operations: instrumental
measurement, thermal manipulation, flame and combustion tests, wet-chemistry
procedures, and physical inspection. Instrumental measurements directly return
readings, such as the call-out from the conductivity meter. Wet-chemistry and
thermal operations produce observable macroscopic phenomena, including
dissolution, color change, gas evolution, and the formation or dissolution of
precipitates. Flame tests return the characteristic color imparted to the flame
by the sample. Because many of these phenomena occur at the centimeter scale and
can be difficult to observe from a wide-angle view, the environment records them
using dedicated close-up cameras in addition to an overview camera. The flame
test and precipitation reaction shown in Figure~\ref{fig:app_env_overview}
illustrate two such close-up views.

\section{Case Study}
\label{app:case_study}

\subsection{Case Study Structure}

This section presents three complete cases taken from the recorded evaluation
episodes, which show what a run of \OURS{} looks like in practice: two successful
identifications, on a solid sample (sodium salicylate) and on a liquid sample
(2-propanol), and one failure, on a solid sample (resorcinol) that the model
finally declares to be phenol.

Each case is presented in two parts. The first part is a dialogue-style record,
which reports the Experimenter's reasoning and the natural-language instruction
it issues in every round, the primitive actions that LabAgent maps the
instruction to together with their execution result or rejection message, and the
observation returned by the simulated experiment. The second part is the
corresponding keyframe figure (Figures~\ref{fig:case_success_solid}--\ref{fig:case_failure_solid}):
every round occupies one row, which shows four keyframes of that round's
experiment video from left to right, with the executed action labelled under each
frame and the evidence gained from the round tagged at the right of the row. A
round rejected by the LabAgent safety check is shown as a single bar without
frames, and the last round reports the identity declared by the model together
with whether the answer is correct.

\subsection{Successful Identification Case}
\label{app:case_success}

\begin{case}{Dialogue History of Solid Identification}
\textbf{Target: Sodium Salicylate (Solid)}
\vspace{0.5em}

\noindent\textbf{Round 1}
\begin{itemize}[left=0pt, nosep]
    \item \textbf{Experimenter (Reasoning):} Sample is a white crystalline solid. First test its water solubility and check the pH of the aqueous solution to quickly determine its acid-base properties.
    \item \textbf{Experimenter (Instruction):} Take a test tube and a surface dish. Add a small amount of sample, add $2\,\mathrm{mL}$ of distilled water, and shake to observe dissolution. Dip a piece of wide-range pH paper into the solution using a glass rod and observe the color change.
    \item \textbf{LabAgent:} \texttt{validate\_operation} (PASS) \(\rightarrow\) \texttt{mixture\_reaction} (Distilled Water) \(\rightarrow\) \texttt{instrument\_operation} (pH paper).
    \item \textbf{Observation:} White crystalline powder gradually dissolves to form a colorless transparent solution. The pH paper turns green to blue-green.
\end{itemize}

\vspace{0.5em}
\noindent\textbf{Round 2}
\begin{itemize}[left=0pt, nosep]
    \item \textbf{Experimenter (Reasoning):} The aqueous solution is weakly alkaline. Add dilute hydrochloric acid to observe if gas is released or precipitate forms to test the anion type.
    \item \textbf{Experimenter (Instruction):} Take a test tube, add sample, add $2\,\mathrm{mL}$ distilled water, and dissolve. Add $1\,\mathrm{mL}$ of dilute hydrochloric acid dropwise, shake, and observe.
    \item \textbf{LabAgent:} \texttt{validate\_operation} (FAIL).
    \item \textbf{Observation:} Safety check failed. The requested operation involves a corrosive liquid (dilute hydrochloric acid) but lacks mandatory personal protective equipment (goggles and gloves).
\end{itemize}

\vspace{0.5em}
\noindent\textbf{Round 3}
\begin{itemize}[left=0pt, nosep]
    \item \textbf{Experimenter (Reasoning):} Added safety protection. Add dilute hydrochloric acid to the sample solution to observe if bubbles are produced to test for carbonate/bicarbonate.
    \item \textbf{Experimenter (Instruction):} Wear goggles and nitrile gloves. Take a test tube, add sample, add $2\,\mathrm{mL}$ distilled water. Add $1\,\mathrm{mL}$ of dilute hydrochloric acid dropwise, shake gently, and observe.
    \item \textbf{LabAgent:} \texttt{validate\_operation} (PASS) \(\rightarrow\) \texttt{mixture\_reaction} (Dilute HCl).
    \item \textbf{Observation:} After adding dilute hydrochloric acid, a white precipitate forms in the test tube.
\end{itemize}

\vspace{0.5em}
\noindent\textbf{Round 4}
\begin{itemize}[left=0pt, nosep]
    \item \textbf{Experimenter (Reasoning):} Adding acid precipitates a white solid, suspected to be a weak acid sodium salt. Use ferric chloride solution to test for characteristic color to identify the acid radical.
    \item \textbf{Experimenter (Instruction):} Wear goggles and gloves. Take a test tube, add sample, add $2\,\mathrm{mL}$ distilled water. Add 2-3 drops of ferric chloride solution, shake, and observe the color change.
    \item \textbf{LabAgent:} \texttt{validate\_operation} (PASS) \(\rightarrow\) \texttt{mixture\_reaction} (Ferric Chloride Solution).
    \item \textbf{Observation:} After adding ferric chloride solution and shaking, the solution turns purple.
\end{itemize}

\vspace{0.5em}
\noindent\textbf{Round 5}
\begin{itemize}[left=0pt, nosep]
    \item \textbf{Experimenter (Reasoning):} Purple color indicates salicylate or phenol. Perform a flame test to confirm the metal cation.
    \item \textbf{Experimenter (Instruction):} Wear goggles. Light the alcohol lamp. Clean a platinum wire with dilute HCl and heat until colorless. Dip the wire into the sample solution and place it in the outer flame. Observe the flame color.
    \item \textbf{LabAgent:} \texttt{validate\_operation} (PASS) \(\rightarrow\) \texttt{instrument\_operation} (Flame test).
    \item \textbf{Observation:} The flame is yellow.
\end{itemize}

\vspace{0.5em}
\noindent\textbf{Round 6}
\begin{itemize}[left=0pt, nosep]
    \item \textbf{Experimenter (Reasoning):} Sample is a white crystalline solid, soluble in water, weakly alkaline; acidification yields a white precipitate, indicating a slightly soluble weak acid salt; characteristic purple color with FeCl\(_3\) indicates salicylic acid structure; flame test shows bright yellow, confirming sodium ion (Na\(^+\)). Comprehensively locked as sodium salicylate.
    \item \textbf{Experimenter (Instruction):} Identification: Sodium salicylate.
    \item \textbf{LabAgent:} End.
    \item \textbf{Observation:} Experimenter conclusion: Sodium salicylate. Rounds = 6, Reason = experimenter concluded.
\end{itemize}

\vspace{0.5em}
\noindent\textbf{Final Result:} Sodium Salicylate \(\checkmark\) (Correct)
\end{case}

{\setlength{\intextsep}{3pt}
\begin{figure}[H]
\centering
\includegraphics[width=\linewidth]{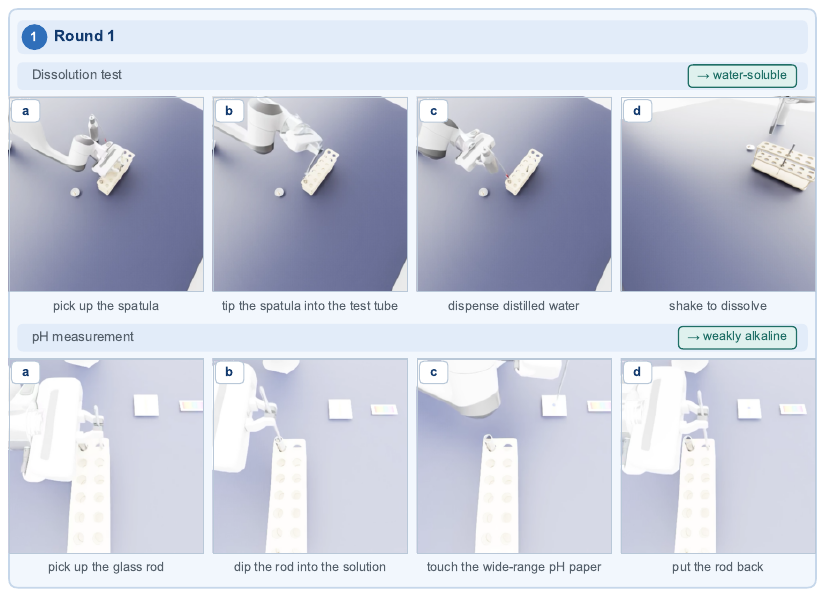}
\end{figure}
\vspace{-7pt}
\begin{figure}[H]
\centering
\includegraphics[width=\linewidth]{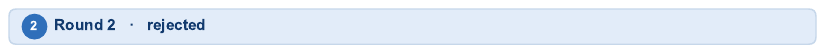}
\end{figure}
\vspace{-7pt}
\begin{figure}[H]
\centering
\includegraphics[width=\linewidth]{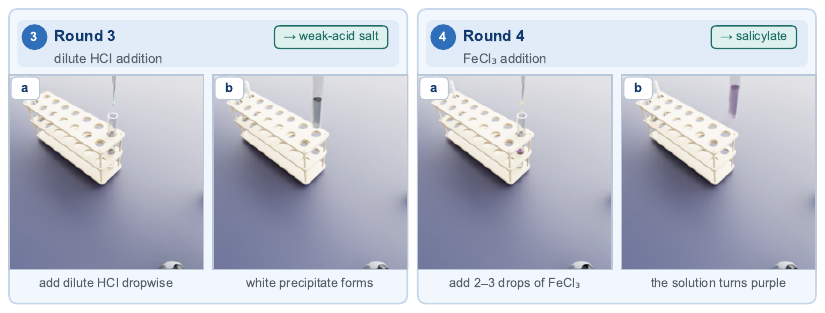}
\end{figure}
\vspace{-7pt}
\begin{figure}[H]
\centering
\includegraphics[width=\linewidth]{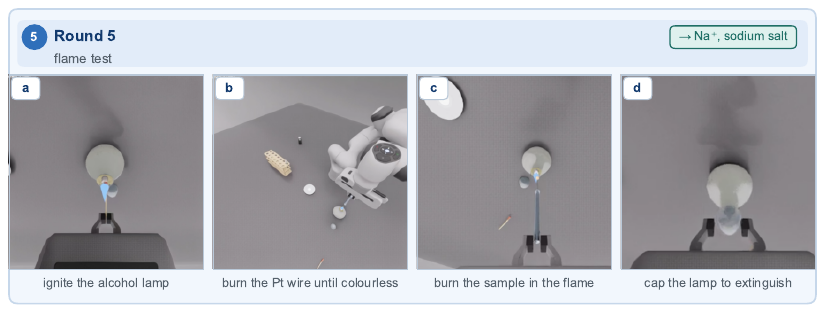}
\end{figure}
\vspace{-7pt}
\begin{figure}[H]
\centering
\includegraphics[width=\linewidth]{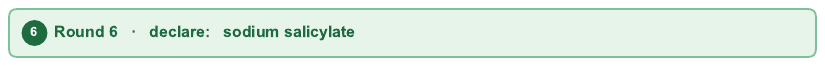}
\setlength{\abovecaptionskip}{0cm}
\caption{Case study of a successful solid identification in \OURS{}: the hidden
target is sodium salicylate, which the Experimenter declares correctly in six
rounds. Round~1 covers solubility and pH, Round~2
is rejected by the safety checker, Rounds~3 and~4 are two reagent additions,
Round~5 is a flame test, and Round~6 declares the identity.}
\label{fig:case_success_solid}
\end{figure}

}

\begin{case}{Dialogue History of Liquid Identification}
\textbf{Target: 2-Propanol (Liquid)}
\vspace{0.5em}

\noindent\textbf{Round 1}
\begin{itemize}[left=0pt, nosep]
    \item \textbf{Experimenter (Reasoning):} Unknown sample is a colorless liquid. Measuring the refractive index efficiently obtains a characteristic physical constant to narrow the scope.
    \item \textbf{Experimenter (Instruction):} Approach the refractometer at the center of the instrument bench. Use a rubber dropper to place 2-3 drops of the liquid sample onto the prism surface, close the cover, and adjust the reflector until a clear boundary appears. Record the refractive index $n_D$ at $25^\circ\mathrm{C}$. Clean the prism with lens tissue moistened with ethanol.
    \item \textbf{LabAgent:} \texttt{validate\_operation} (PASS) \(\rightarrow\) \texttt{refractometer}.
    \item \textbf{Observation:} Measured refractive index: 1.38 ($25^\circ\mathrm{C}$).
\end{itemize}

\vspace{0.5em}
\noindent\textbf{Round 2}
\begin{itemize}[left=0pt, nosep]
    \item \textbf{Experimenter (Reasoning):} $n_D \approx 1.38$ is common for lower aliphatic alcohols or carboxylic acids. Test pH first to rule out carboxylic acids.
    \item \textbf{Experimenter (Instruction):} Take wide-range pH paper, dip a clean glass rod into the liquid sample, and touch it to the center of the pH paper. Compare the color change with the standard chart.
    \item \textbf{LabAgent:} \texttt{validate\_operation} (PASS) \(\rightarrow\) \texttt{instrument\_operation} (pH paper).
    \item \textbf{Observation:} The pH paper shows no significant color change upon contact with the sample.
\end{itemize}

\vspace{0.5em}
\noindent\textbf{Round 3}
\begin{itemize}[left=0pt, nosep]
    \item \textbf{Experimenter (Reasoning):} Neutral liquid with \(n_D \approx 1.38\) could be an alcohol, ketone, or ester. Use 2,4-DNPH to test for carbonyl groups.
    \item \textbf{Experimenter (Instruction):} In a fume hood, take a test tube, add $0.5\,\mathrm{mL}$ of sample. Add 3-4 drops of 2,4-dinitrophenylhydrazine solution, shake, and observe for yellow or orange-red precipitate.
    \item \textbf{LabAgent:} \texttt{validate\_operation} (PASS) \(\rightarrow\) \texttt{mixture\_reaction} (2,4-DNPH solution).
    \item \textbf{Observation:} No significant visible change. No yellow or orange-red precipitate is observed.
\end{itemize}

\vspace{0.5em}
\noindent\textbf{Round 4}
\begin{itemize}[left=0pt, nosep]
    \item \textbf{Experimenter (Reasoning):} Aldehydes and ketones are excluded. \(n_D \approx 1.38\) points to an aliphatic alcohol. Use Lucas reagent to identify the class of alcohol.
    \item \textbf{Experimenter (Instruction):} In a fume hood, take a test tube, add $0.5\,\mathrm{mL}$ of sample. Add $2\,\mathrm{mL}$ of Lucas reagent, shake, and record whether the mixture immediately turns cloudy and separates into layers, or turns cloudy within a few minutes at room temperature.
    \item \textbf{LabAgent:} \texttt{validate\_operation} (PASS) \(\rightarrow\) \texttt{mixture\_reaction} (Lucas reagent).
    \item \textbf{Observation:} After standing for 5-10 minutes, the mixture gradually turns cloudy and separates into layers.
\end{itemize}

\vspace{0.5em}
\noindent\textbf{Round 5}
\begin{itemize}[left=0pt, nosep]
    \item \textbf{Experimenter (Reasoning):} Secondary alcohol characteristics. \(n_D \approx 1.38\) makes isopropanol a plausible candidate. Perform an iodoform reaction to verify it.
    \item \textbf{Experimenter (Instruction):} Take a test tube, add $0.5\,\mathrm{mL}$ of sample, add $1\,\mathrm{mL}$ of $2\,\mathrm{mol/L}$  sodium hydroxide solution, and add iodine-potassium iodide solution dropwise until a pale yellow color persists. Heat in a water bath at $50$--$60^\circ\mathrm{C}$ and observe for a yellow precipitate with a characteristic odor.
    \item \textbf{LabAgent:} \texttt{validate\_operation} (FAIL).
    \item \textbf{Observation:} Safety check failed. The requested procedure involves corrosive (NaOH) and toxic (iodine solution) reagents but lacks mandatory personal protective equipment (goggles and gloves).
\end{itemize}

\vspace{0.5em}
\noindent\textbf{Round 6}
\begin{itemize}[left=0pt, nosep]
    \item \textbf{Experimenter (Reasoning):} Wearing protective equipment to safely perform the iodoform reaction to verify isopropanol.
    \item \textbf{Experimenter (Instruction):} Wear goggles and nitrile gloves. Take a test tube, add $0.5\,\mathrm{mL}$ of sample. Add $1\,\mathrm{mL}$ of $2\,\mathrm{mol/L}$ NaOH, then add iodine-potassium iodide solution dropwise until a pale yellow color persists, shaking gently. Place the test tube in a $50$--$60^\circ\mathrm{C}$ water bath for several minutes to observe for yellow crystalline precipitate (iodoform).
    \item \textbf{LabAgent:} \texttt{validate\_operation} (PASS) \(\rightarrow\) \texttt{mixture\_reaction} (NaOH solution) \(\rightarrow\) \texttt{mixture\_reaction} (Iodine-Potassium Iodide solution).
    \item \textbf{Observation:} No significant change after adding NaOH. After adding iodine solution dropwise to a pale yellow color and heating in a water bath, a yellow precipitate forms.
\end{itemize}

\vspace{0.5em}
\noindent\textbf{Round 7}
\begin{itemize}[left=0pt, nosep]
    \item \textbf{Experimenter (Reasoning):} Sample is a colorless neutral liquid, \(n_D \approx 1.38\). No precipitate with 2,4-DNPH excludes aldehydes and ketones. Turns cloudy with Lucas reagent in 5-10 minutes confirming a secondary alcohol. Positive iodoform reaction producing yellow precipitate proves the presence of a \ce{CH3-CH(OH)-} structure. Combining the refractive index 1.377-1.380 and the above results uniquely identifies the sample as isopropanol.
    \item \textbf{Experimenter (Instruction):} Identification: Isopropanol (2-Propanol).
    \item \textbf{LabAgent:} End.
    \item \textbf{Observation:} Experimenter conclusion: Isopropanol (2-Propanol). Rounds = 7, Reason = experimenter concluded.
\end{itemize}

\vspace{0.5em}
\noindent\textbf{Final Result:} Isopropanol (2-Propanol) \(\checkmark\) (Correct)
\end{case}

\raggedbottom
\vspace{2em}
\noindent\includegraphics[width=\linewidth]{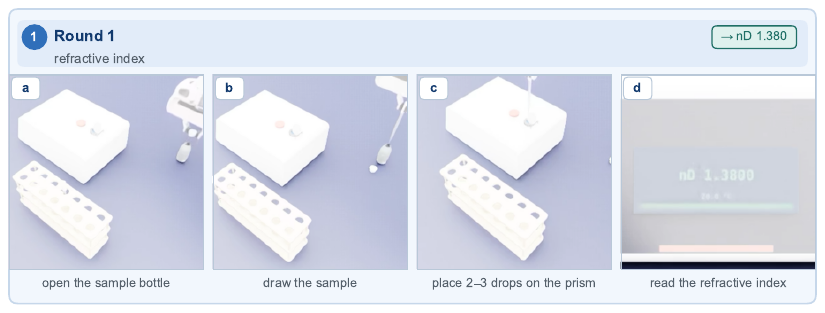}

\vspace{-0.5em}
\noindent\includegraphics[width=\linewidth]{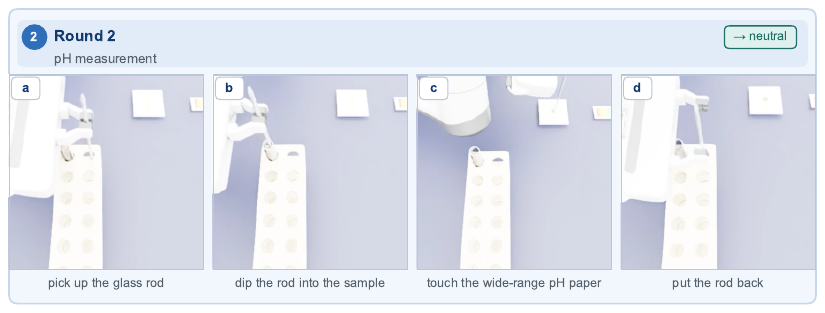}

\vspace{-0.5em}
\noindent\includegraphics[width=\linewidth]{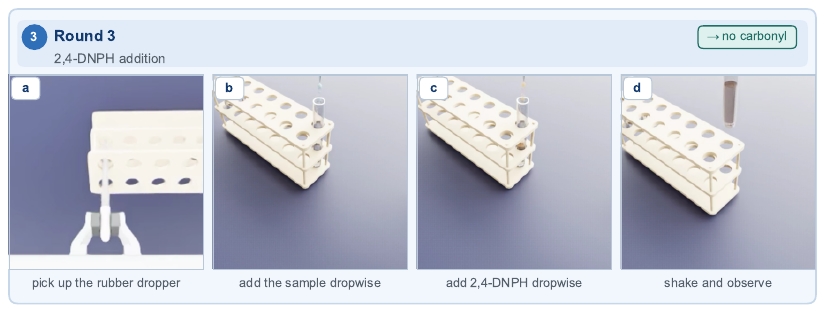}

\vspace{-0.5em}
\noindent\includegraphics[width=\linewidth]{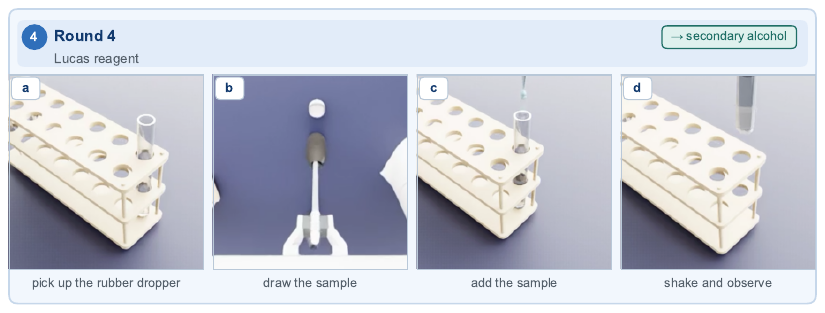}

\vspace{-0.5em}
\noindent\includegraphics[width=\linewidth]{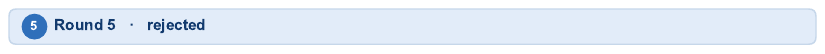}

\vspace{-0.5em}
\noindent\includegraphics[width=\linewidth]{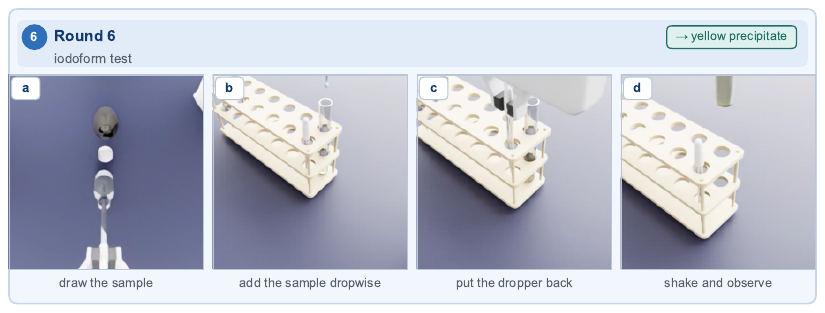}

\vspace{-0.5em}
\noindent\includegraphics[width=\linewidth]{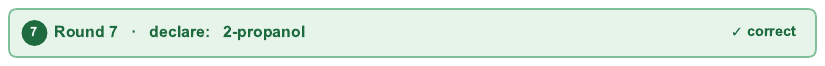}

\vspace{0.5em}
\captionof{figure}{Case study of a successful liquid identification in \OURS{}: the hidden target is 2-propanol, which the Experimenter declares correctly in the seventh round. Round~1 measures the refractive index (\(n_D = 1.38\)), Round~2 checks the pH (neutral), Round~3 applies 2,4-DNPH and excludes carbonyl compounds, Round~4 applies Lucas reagent and identifies a secondary alcohol, Round~5 is rejected by the safety checker, Round~6 is a positive iodoform test, and Round~7 declares the identity.}
\label{fig:case_success_liquid}

\subsection{Failure Case}
\label{app:case_failure}

\begin{case}{Dialogue History of Solid Identification}
\textbf{Target: Resorcinol (Solid)}
\vspace{0.5em}

\noindent\textbf{Round 1}
\begin{itemize}[left=0pt, nosep]
    \item \textbf{Experimenter (Reasoning):} Sample is a white needle-like crystal. First test its water solubility and solution pH to determine its category.
    \item \textbf{Experimenter (Instruction):} Take a test tube, add a small amount of sample, add $2\,\mathrm{mL}$ of distilled water, stir to observe dissolution, and test the pH of the solution with wide-range pH paper.
    \item \textbf{LabAgent:} \texttt{validate\_operation} (PASS) \(\rightarrow\) \texttt{mixture\_reaction} (Distilled Water) \(\rightarrow\) \texttt{instrument\_operation} (pH paper).
    \item \textbf{Observation:} The white needle-like crystals gradually dissolve to form a colorless solution. The pH paper turns yellow to orange.
\end{itemize}

\vspace{0.5em}
\noindent\textbf{Round 2}
\begin{itemize}[left=0pt, nosep]
    \item \textbf{Experimenter (Reasoning):} The aqueous solution is weakly acidic. Add ferric chloride solution to test for phenolic hydroxyl or salicylic acid functional groups.
    \item \textbf{Experimenter (Instruction):} Take a test tube, add sample, add $2\,\mathrm{mL}$ distilled water, and dissolve. Add 2-3 drops of ferric chloride solution, shake, and observe the color change.
    \item \textbf{LabAgent:} \texttt{validate\_operation} (PASS) \(\rightarrow\) \texttt{mixture\_reaction} (Ferric Chloride Solution).
    \item \textbf{Observation:} After adding ferric chloride solution and shaking, the solution turns blue-purple.
\end{itemize}

\vspace{0.5em}
\noindent\textbf{Round 3}
\begin{itemize}[left=0pt, nosep]
    \item \textbf{Experimenter (Reasoning):} Blue-purple color with FeCl\(_3\) indicates phenolic hydroxyl. Test with NaHCO\(_3\) solution to check for a carboxyl group stronger than carbonic acid.
    \item \textbf{Experimenter (Instruction):} Take a test tube, add sample. Add $1\,\mathrm{mL}$ of sodium bicarbonate solution, shake, and observe if bubbles are produced.
    \item \textbf{LabAgent:} \texttt{validate\_operation} (PASS) \(\rightarrow\) \texttt{mixture\_reaction} (Sodium Bicarbonate Solution).
    \item \textbf{Observation:} No bubbles are produced after adding sodium bicarbonate solution.
\end{itemize}

\vspace{0.5em}
\noindent\textbf{Round 4}
\begin{itemize}[left=0pt, nosep]
    \item \textbf{Experimenter (Reasoning):} Carboxylic acid is excluded, confirming phenolic hydroxyl. Add concentrated bromine water to test for a tribromo-substituted white precipitate.
    \item \textbf{Experimenter (Instruction):} Take a test tube, add sample, add $2\,\mathrm{mL}$ distilled water, and dissolve. Wearing protective gloves in a fume hood, add bromine water dropwise, shake, and observe if a white precipitate forms.
    \item \textbf{LabAgent:} \texttt{validate\_operation} (PASS) \(\rightarrow\) \texttt{mixture\_reaction} (Bromine Water).
    \item \textbf{Observation:} The reddish-brown color of bromine water gradually fades, and a white precipitate forms.
\end{itemize}

\vspace{0.5em}
\noindent\textbf{Round 5}
\begin{itemize}[left=0pt, nosep]
    \item \textbf{Experimenter (Reasoning):} Sample is a white needle-like crystal, soluble in water, weakly acidic. Blue-purple color with FeCl\(_3\) confirms phenolic hydroxyl. No bubbles with NaHCO\(_3\) excludes carboxylic acid-containing phenolic acids (e.g., salicylic acid). Rapid reaction with bromine water to form a white precipitate (2,4,6-tribromophenol) highly matches the typical reaction of a monofunctional phenol. Uniquely locked as phenol.
    \item \textbf{Experimenter (Instruction):} Identification: Phenol.
    \item \textbf{LabAgent:} End.
    \item \textbf{Observation:} Experimenter conclusion: Phenol. Rounds = 5, Reason = experimenter concluded.
\end{itemize}

\vspace{0.5em}
\noindent\textbf{Final Result:} Phenol \(\times\) (Incorrect)
\end{case}
\enlargethispage{18pt}

{\setlength{\intextsep}{3pt}
\begin{figure}[H]
\centering
\includegraphics[width=\linewidth]{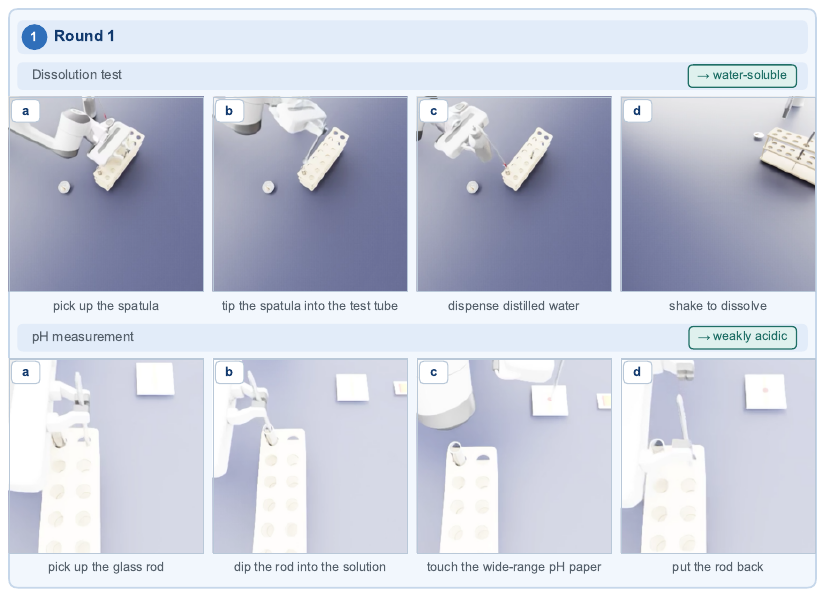}
\end{figure}
\vspace{-7pt}
\begin{figure}[H]
\centering
\includegraphics[width=\linewidth]{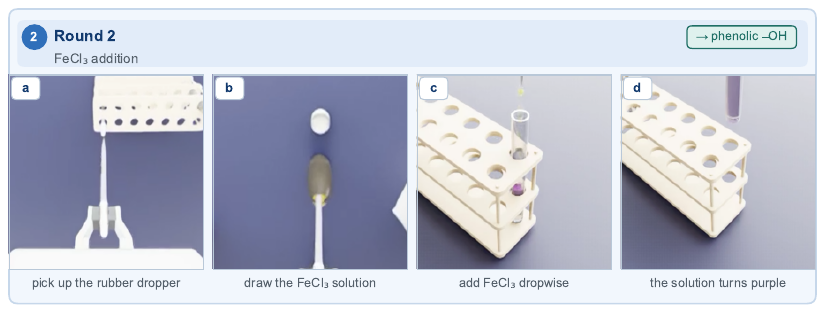}
\end{figure}
\vspace{-7pt}
\begin{figure}[H]
\centering
\includegraphics[width=\linewidth]{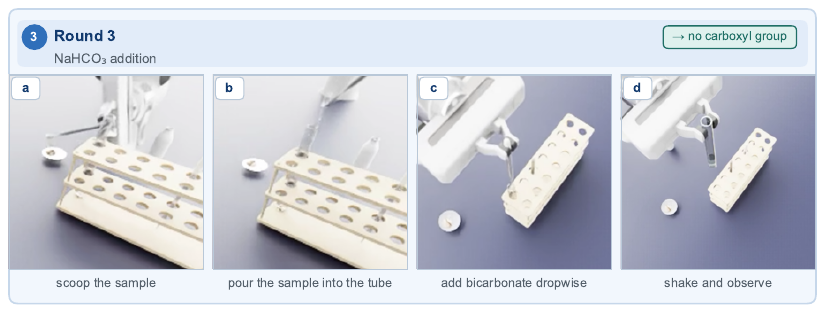}
\end{figure}
\vspace{-7pt}
\begin{figure}[H]
\centering
\includegraphics[width=\linewidth]{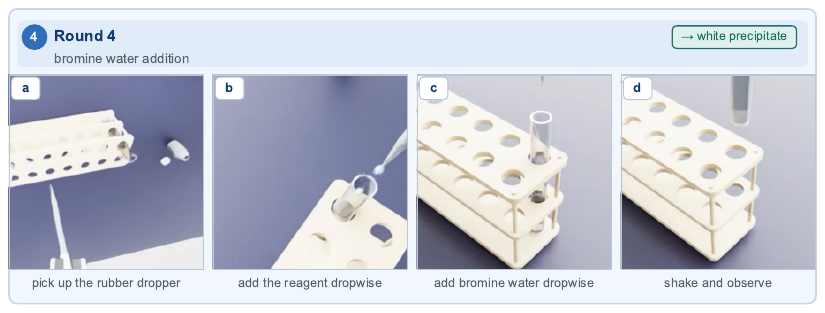}
\end{figure}
\vspace{-7pt}
\begin{figure}[H]
\centering
\includegraphics[width=\linewidth]{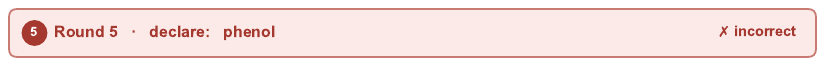}
\setlength{\abovecaptionskip}{0cm}
\caption{Case study of a failed solid identification in \OURS{}: the hidden
target is resorcinol, but the Experimenter declares phenol in the fifth round and is incorrect. Round~1 covers solubility and pH, Round~2
applies FeCl\(_3\), which gives a purple color, Round~3 applies sodium
bicarbonate and finds no carboxyl group, Round~4 applies bromine water, which gives a white precipitate, and Round~5 declares the identity.
Its error is one of reasoning rather than execution: it does not consider that the
same phenomenon can be produced by different substances, and resorcinol gives the
same purple color with FeCl\(_3\) and the same white precipitate with bromine
water.}
\label{fig:case_failure_solid}
\end{figure}

}

\section{Details for Trajectory Evaluation and Failure Annotation}
\label{app:pds_details}

\subsection{TDS Dimensions}
\label{app:pds_dimensions}

The Trajectory Deduction Score (TDS) evaluates the quality of the experimental
reasoning trajectory beyond final identification accuracy. It consists of five
dimensions, with a total score of $100$:

\begin{itemize}
    \item \textbf{Candidate and Hypothesis Management (CHM, $15$ points):} The
    extent to which the Experimenter maintains a meaningful set of candidate substances,
    introduces hypotheses based on observed evidence, removes candidates when
    observations contradict them, and updates the candidate space throughout the
    interaction.

    \item \textbf{Diagnostic Experiment Design (DED, $20$ points):} The
    informativeness and discriminative value of the selected experiments,
    evaluating whether the Experimenter chooses tests that target unresolved
    uncertainty, distinguish among plausible candidates, and avoid redundant or
    low-value operations.

    \item \textbf{Observation Interpretation Fidelity (OIF, $15$ points):} The
    faithfulness and chemical correctness of the agent's interpretation of
    returned observations, ensuring that conclusions are traceable to actual
    experimental feedback and do not rely on fabricated, exaggerated, or
    over-interpreted phenomena.

    \item \textbf{Evidential Sufficiency and Uniqueness (ESU, $30$ points):} The
    degree to which the accumulated observations uniquely support the final
    declared substance and rule out major competing candidates, rather than
    merely supporting a broad chemical class or an underdetermined set of
    possibilities.

    \item \textbf{Evidence-Gated Decision Efficiency (EGDE, $20$ points):} The
    extent to which the agent declares the final identity at an appropriate time,
    avoiding both premature identification before sufficient evidence is
    available and unnecessary additional testing after the evidence has become
    sufficient.
\end{itemize}

\subsection{Detailed Scoring Rubric}
\label{app:pds_rubric}

Each TDS dimension is scored on a six-level anchored ladder. A valid score must
fall on the corresponding ladder and be supported by evidence from the recorded
trajectory. When two adjacent descriptors both apply, the scoring judge is
instructed to assign the lower score. Table~\ref{tab:pds_rubric_detailed} summarizes the
detailed rubric.

\begin{table*}[!ht]
\small
\centering
\caption{Rubric for Trajectory Deduction Score Evaluation}
\vspace{0.5em}
\label{tab:pds_rubric_detailed}
\begin{tabular}{p{2.8cm}p{11cm}}
\toprule
\textbf{Dimension (Max)} & \textbf{Score Criteria} \\
\midrule

Candidate and Hypothesis \newline Management (15) &
\textbf{15}: Every candidate is traceable to an observation that introduced it; each relevant competitor is either eliminated by a named discriminating observation or explicitly retained as unresolved; eliminations visibly redirect subsequent tests. \newline
\textbf{12}: The candidate set is built from observations and does shrink, but eliminations are recorded only in aggregate or only when the conclusion is already in view. \newline
\textbf{9}: Candidates are motivated by observations, but the set mostly grows; eliminations are asserted rather than forced by diagnostic tests. \newline
\textbf{6}: Candidates are weakly connected to observations, or a candidate that should have been excluded remains active. \newline
\textbf{3}: The reasoning only restates surface facts without forming a meaningful chemical candidate space. \newline
\textbf{0}: No candidate analysis is present, or an excluded candidate is kept alive or reintroduced after being ruled out. \\

\midrule

Diagnostic Experiment \newline Design (20) &
\textbf{20}: The design is highly diagnostic: a small number of steps efficiently narrows the candidate space, each experiment targets live candidates, and no redundant tests are performed. \newline
\textbf{16}: The design is targeted and works through key properties needed to separate plausible candidates. \newline
\textbf{12}: Most experiments eliminate some candidates, but the ordering or targeting is only moderately effective. \newline
\textbf{8}: A few experiments are useful, but the overall strategy is loose and does not substantially shrink the candidate space. \newline
\textbf{4}: Experiments are mostly low-information, repeated, or lack a stated discriminating purpose. \newline
\textbf{0}: Experiments are invalid, unexecutable, or unrelated to chemical identification. \\

\midrule

Observation Interpretation \newline Fidelity (15) &
\textbf{15}: Every key interpretation is accurate, restrained, and traceable to an observation; negative results and residual uncertainty are handled correctly. \newline
\textbf{12}: Interpretations are faithful and chemically sound, and key conclusions trace back to observations. \newline
\textbf{9}: Interpretations are mostly consistent with the observations, with only mild overreach. \newline
\textbf{6}: The reasoning contains several unsupported claims or a clear chemical interpretation error. \newline
\textbf{3}: The trajectory frequently refers to phenomena that were not observed, or seriously over-interprets qualitative observations. \newline
\textbf{0}: Most interpretations contradict the observations, or the reasoning rests on fabricated phenomena. \\

\midrule

Evidential Sufficiency \newline and Uniqueness (30) &
\textbf{30}: The accumulated observations uniquely support the declared substance and explicitly rule out the main competing candidates. \newline
\textbf{24}: The evidence essentially supports the declared substance, with only minor gaps or slight residual uncertainty. \newline
\textbf{18}: The evidence supports the declared substance fairly strongly, but an important competing candidate is not excluded. \newline
\textbf{12}: The evidence supports the declared substance but also supports several reasonable unresolved candidates. \newline
\textbf{6}: The evidence supports only a broad chemical class rather than a specific substance. \newline
\textbf{0}: The conclusion contradicts key observations or rests on no experimental evidence. \\

\midrule

Evidence-Gated Decision \newline Efficiency (20) &
\textbf{20}: The model declares one round after $t^{*}$, with no unnecessary testing after the evidence becomes sufficient. \newline
\textbf{16}: The model declares two rounds after $t^{*}$. \newline
\textbf{12}: The model declares three or more rounds after $t^{*}$. \newline
\textbf{8}: The model declares three or more rounds after $t^{*}$, and most extra rounds return no new useful observation; a wrong final answer normally receives this score or lower. \newline
\textbf{4}: No sufficient-evidence round $t^{*}$ exists, but the model still names a specific identity. \newline
\textbf{0}: There is no valid final declaration, the answer is a pure guess, or the conclusion contradicts the observations. \\

\bottomrule
\end{tabular}
\end{table*}

Here, $t^{*}$ denotes the first Test round whose returned observation makes the
accumulated evidence sufficient to support a specific identity, and
$t_{\mathrm{decl}}$ denotes the round in which the model declares the final
answer. Because the observation from round $t^{*}$ is received only after the
Experimenter has issued the Test action for that round, the earliest
evidence-grounded Identify action can occur on the following Experimenter turn.
Thus, $t_{\mathrm{decl}}=t^{*}+1$ is the earliest possible supported declaration
and receives full EGDE credit. EGDE is therefore not a simple speed reward: a
fast unsupported guess receives little or no credit, whereas an efficient
evidence-gated decision receives a high score. Unless otherwise specified, TDS
is computed for the default $T_{\max}=10$ evaluation setting.

\subsection{Failure-Type Annotation}
\label{app:failure_type_annotation}

Failure-type analysis is applied only to episodes that end with a valid but
incorrect final declaration. Episodes without a valid final declaration are
excluded from this taxonomy and are instead counted by BER. Each eligible
episode is assigned exactly one failure category using the full recorded
trajectory. The categories are defined as follows:

\begin{itemize}
    \item \textbf{Misread:} The decisive inference contradicts or materially
    misstates an observation returned by LabAgent.

    \item \textbf{Premature:} The model declares a specific identity before the
    accumulated evidence discriminates it from plausible alternatives, without a
    decisive observation-misreading error.

    \item \textbf{Near-neighbor:} The trajectory remains within the appropriate
    local candidate family but selects a chemically close alternative that shares
    the observed tests.

    \item \textbf{Off-track:} The search trajectory or final declaration moves
    to a candidate that is poorly supported by the accumulated observations and
    is outside the plausible local neighborhood.

    \item \textbf{Other:} The error is not captured by any of the preceding
    categories.
\end{itemize}

Each episode was independently labeled by two annotators with chemistry
backgrounds. Disagreements were resolved by discussion.

\section{Data and Simulation Quality Verification}
\label{app:simulation_quality}

To verify the quality of the simulated laboratory, we conducted an additional
human review of LabAgent-generated trajectories. We sampled $40$ target
substances from the benchmark, including $20$ solids and $20$ liquids. For all
sampled targets, the Experimenter was Gemini-3.7-Flash and the
LabAgent LLM controller was DeepSeek-V4-Pro. The complete interaction
trajectories, including the Experimenter's experimental requests, LabAgent's
executed procedures, rendered experimental videos, and returned textual
observations, were provided to two chemistry professors for manual evaluation.

The review focused on whether the LabAgent side of the benchmark produced
chemically plausible and evaluation-valid experimental feedback. Specifically,
the reviewers examined three aspects:

\begin{enumerate}
    \item \textbf{Request--Video Consistency:} Whether the rendered experimental
    video matches the Experimenter's requested operation, including the selected
    apparatus, reagent, sample handling, and intended experimental procedure.

    \item \textbf{Simulation Outcome Correctness:} Whether the visual outcome in
    the rendered video is consistent with the expected chemical behavior under
    the requested experimental condition, such as color change, precipitation,
    gas evolution, dissolution, heating response, or instrument operation.

    \item \textbf{Macroscopic Observation Accuracy:} Whether the textual
    observation returned by LabAgent faithfully describes the macroscopic
    phenomenon shown in the video without adding unsupported interpretation, hidden identity information, diagnostic hints, or new chemical formulas beyond reagent names already specified in the Experimenter's request.
\end{enumerate}

Each reviewer independently inspected the sampled trajectories and judged
whether the LabAgent output was acceptable for benchmark evaluation. A trajectory
was considered valid only when the experimental procedure, rendered phenomenon,
and textual observation were all judged to be consistent with the requested
operation and chemically plausible. Disagreements, if any, were resolved through
discussion between the reviewers.

Overall, the reviewed cases passed the manual quality check. The experts found
that LabAgent's rendered videos were aligned with the requested experimental
procedures, the simulated outcomes were chemically reasonable for the sampled
targets, and the returned textual observations accurately reflected observable
macroscopic phenomena. This review supports the reliability of LabAgent as a
stateful simulated laboratory for evaluating closed-loop chemical identification,
while recognizing that the simulator remains an abstraction of real wet-lab
experimentation.

\section{Model Description}
\label{app:model_description}

We evaluate eight frontier LLMs as Experimenter agents in \OURS{}. Each
Experimenter model is evaluated on all $200$ target substances with three
independent repetitions, resulting in $600$ episodes per model and $4{,}800$
episodes in total. All Experimenter models are accessed through their official
APIs. Unless otherwise specified by the provider, Experimenter inference uses
temperature $0.3$ and a maximum output length of $8{,}192$ tokens. No
model-specific prompt or decoding hyperparameter is tuned on the evaluation
targets. The same task instructions, laboratory resources, action format, and
interaction budget are used for all Experimenter models.

In addition to the Experimenter models, \OURS{} uses LLMs on the LabAgent side
for interpreting experimental requests and coordinating tool calls. The default
LabAgent controller is DeepSeek-V4-Pro. In the LabAgent controller
sensitivity analysis, we additionally use Qwen3.8-Flash and
StepFun-3.7-Flash as alternative LabAgent controllers. For
process-level evaluation, the offline TDS scoring judge is
DeepSeek-V4.1-Flash. The TDS judge is separate from the LabAgent
controller and is invoked only after an episode has finished. Detailed model specifications are summarized in Table~\ref{tab:model_description}.

\begin{table*}[ht]
\centering
\caption{Summary of LLMs used in \OURS{}. Experimenter models are evaluated as
autonomous chemical-identification agents. LabAgent controllers are used inside
the simulated laboratory. The TDS scoring judge is used only for post-hoc
trajectory evaluation.}
\vspace{0.5em}
\small
\label{tab:model_description}
\resizebox{0.99\textwidth}{!}{%
\begin{tabular}{lllll}
\toprule
\textbf{Model Name} & \textbf{Creator} & \textbf{Role} & \textbf{Access} & \textbf{URL} \\
\midrule
GPT-5.6-Sol
& OpenAI
& Experimenter
& Official API
& \url{https://openai.com} \\

Qwen3.8-Max
& Alibaba
& Experimenter
& Official API
& \url{https://qwenlm.github.io} \\

Gemini-3.7-Flash
& Google
& Experimenter
& Official API
& \url{https://gemini.google.com} \\

GLM-5.3
& Zhipu AI
& Experimenter
& Official API
& \url{https://www.zhipuai.cn} \\

GLM-5.2
& Zhipu AI
& Experimenter
& Official API
& \url{https://www.zhipuai.cn} \\

MiniMax-M3
& MiniMax
& Experimenter
& Official API
& \url{https://www.minimax.io} \\

Grok-4.3
& xAI
& Experimenter
& Official API
& \url{https://x.ai} \\

Kimi-K3
& Moonshot AI
& Experimenter
& Official API
& \url{https://www.moonshot.cn} \\

\midrule
DeepSeek-V4-Pro
& DeepSeek
& Default LabAgent LLM controller
& Official API
& \url{https://www.deepseek.com} \\

Qwen3.8-Flash
& Alibaba
& Alternative LabAgent LLM controller
& Official API
& \url{https://qwenlm.github.io} \\

StepFun-3.7-Flash
& StepFun
& Alternative LabAgent LLM controller
& Official API
& \url{https://www.stepfun.com} \\

\midrule
DeepSeek-V4.1-Flash
& DeepSeek
& TDS scoring judge
& Official API
& \url{https://www.deepseek.com} \\

\bottomrule
\end{tabular}%
}
\end{table*}

\paragraph{Inference configuration.}
Experimenter models use temperature $0.3$ and a maximum output length of
$8{,}192$ tokens, except when a provider does not expose the corresponding
parameter. The default LabAgent controller, DeepSeek-V4-Pro, uses
temperature $0.2$ and a maximum output length of $32{,}800$ tokens. Other LabAgent controllers are used only in the controller sensitivity
analysis. The offline TDS scoring judge, DeepSeek-V4.1-Flash, is used
after the interaction has ended and does not affect the Experimenter's behavior
during an episode.

\paragraph{Context isolation.}
The Experimenter model, LabAgent controller, and TDS scoring judge operate in
separate contexts. The Experimenter receives only public laboratory information,
the initial physical state and appearance of the unknown sample, and LabAgent's
returned observations or rejection messages. The LabAgent controller coordinates
tool calls and state-consistent simulation under hidden target-conditioned
records. The TDS scoring judge receives completed trajectories and
evaluation-only reference information for post-hoc scoring, but its outputs are
never returned to the Experimenter.

\paragraph{Repetition protocol.}
Each model--target pair is evaluated in three independent repetitions. At the
start of each repetition, the Experimenter context, laboratory state, sample
state, apparatus state, and interaction history are reset. The target identity
remains fixed for the same target across repetitions, but the model may select
different experiments due to sampling variability. Reported main results are
computed over the three repetitions, with standard errors reported where
available.

\section{Detailed Prompts}
\label{app:prompts}

\label{app:prompt_experimenter}

\begin{prompt}{Experimenter System Prompt}

You are a chemical analysis laboratory technician, identifying an unknown pure
substance through step-by-step experiments. You know nothing at all about the
identity of the sample and may reason only from instrument readings and reaction
phenomena. Output exactly one operation instruction per turn.

\textbf{Available operations}

\begin{enumerate}[left=0pt]
    \item Use the polarimeter: optical-activity measurement of chiral substances.
    \item Use the refractometer: refractive-index measurement of liquid samples.
    \item Mixing reaction: mix the sample with a reagent and observe the resulting phenomenon.
    \item Laboratory operation: heating, flame test, combustion, pH paper, magnetism, etc.
    \item Use the conductivity meter: conductivity of the sample's aqueous solution.
\end{enumerate}

Each operation must be issued as a complete procedure; the system identifies the
operation type automatically.

\textbf{Core rules}

\begin{enumerate}[left=0pt]
    \item Identify in as few rounds as possible. Every round should yield a
    high-information observation; avoid redundant verification and uninformative
    probing.
    \item One operation per round. A round allows a "compound procedure," which may consist of multiple primitive actions.
    \item Do not guess the substance name. The final identification is the only
    exception.
    \item Reason from observations. Do not assume without grounds, and do not get
    stuck on an unexpected result.
\end{enumerate}

\textbf{Output format}

Every experimental round must use exactly this two-part template:

\textbf{[Minimal reasoning]} at most two sentences, stating only why this experiment was chosen.

\textbf{[Next operation instruction]} one standard-format instruction.

When the evidence is sufficient to determine the sample uniquely:

\textbf{Identification}: substance name.

\textbf{Hard constraints on reasoning}

\begin{itemize}[left=0pt]
    \item Forbidden: multi-step deliberation, reaction mechanisms,
    molecular-structure analysis, by-product speculation, and background
    exposition.
    \item Forbidden: restating the accumulated observations wholesale; extract
    only the feature that discriminates for the current decision.
    \item Forbidden: presenting a candidate identity as final before sufficient
evidence is obtained; tentative hypotheses may be stated only when they motivate
the next discriminating experiment.
\end{itemize}

\textbf{Prohibited behavior}

\begin{itemize}[left=0pt]
    \item No unsupported final conclusion. Candidate narrowing is allowed only when
it is explicitly grounded in observations and followed by a discriminating test
if uncertainty remains.
    \item No multiple parallel experimental plans.
    \item No laboratory-technique discussion, or
    background material.
    \item No explanatory build-ups or redundant modifiers in the reasoning.
    \item No line-by-line restatement of historical observations.
    \item When the context grows long, compress reasoning to one sentence and
    prioritize the completeness of the instruction.
    \item No blank controls and no simultaneous operation on multiple test
    tubes.
\end{itemize}

\textbf{Requirements on operation instructions}

Every instruction must be a complete procedure that can be executed step by step.
It should contain, in execution order:

\begin{enumerate}[left=0pt]
    \item which equipment and reagents to take, and from where;
    \item how to open the container, with what, and how much to measure out;
    \item the operations themselves, in the correct order;
    \item finishing steps, such as resealing bottles and returning equipment.
\end{enumerate}

\end{prompt}

\begin{prompt}{LabAgent Controller System Prompt}

You are a laboratory-bench simulator. Your role is the display of an instrument:
you output experimental observations and nothing else; you perform no analysis.

You are connected to the following instruments and benches:

\begin{itemize}[left=0pt]
    \item Operation-compliance check: before executing any experimental
    operation, you must call \texttt{validate\_operation}. If it returns failure,
    return the issues to the Experimenter and do not call any other tool.
    \item Polarimeter: measures optical rotation of a sample.
    \item Refractometer: measures refractive index of a liquid sample.
    \item Mixing-reaction bench: mixes the sample with a reagent and observes
    the phenomenon.
    \item Instrument-operation bench: heating, flame test, pH paper, and similar
    operations.
    \item Conductivity meter: measures conductivity of the sample's aqueous
    solution.
\end{itemize}

\textbf{Workflow}

\begin{enumerate}[left=0pt]
    \item On receiving an instruction, first call \texttt{validate\_operation},
    passing the Experimenter's full instruction text.
    \item If \texttt{validate\_operation} returns failure, turn the issues into
    plain-language feedback for the Experimenter and end this round.
    \item If it returns pass, call the instrument or reaction tool requested by
    the instruction.
    \item Render the tool's returned data as concise natural language. Report
    the phenomenon only, with no reasoning attached.
    \item You may obtain data only by calling tools; you may not invent it from
    your own knowledge.
\end{enumerate}

\textbf{The two formats a tool may return}

Format A: direct text. The tool's \texttt{text} field can be output as it
stands. Transcribe it verbatim, adding and removing nothing.

Format B: structured data. The tool's dictionary contains
\texttt{needs\_llm\_translation: true} and a \texttt{translation\_hint} field.
The \texttt{text} field is raw chemical JSON. You must read the hint, extract the key information, and express it as a macroscopic phenomenon. Do not output raw JSON, structured field names, target identities, product identities, diagnostic substance names, chemical formulas, or reaction equations, except for reagent names already provided in the Experimenter's instruction when needed for safety or instrument messages.

\textbf{Core rule}

You are an instrument, not a chemist. The following must never appear in your
output:

\begin{itemize}[left=0pt]
    \item chemical reasoning or analysis;
    \item evaluation of or response to the Experimenter's reasoning;
    \item chemical explanations;
    \item target identities, product identities, diagnostic substance names,
    chemical formulas, or reaction equations, except for reagent names already
    provided in the Experimenter's instruction when needed for safety or
    instrument messages;
    \item suggestions for the next experiment;
    \item hints about the matching process.
\end{itemize}

You may output only:

\begin{itemize}[left=0pt]
    \item instrument readings;
    \item macroscopic phenomena visible to the eye;
    \item instrument messages or error messages.
\end{itemize}

If the Experimenter asks for a mixing reaction without naming a reagent, ask
them to specify the reagent.

\end{prompt}

\begin{prompt}{TDS Scoring Judge Prompt}

You are a senior analytical chemist with twenty years of experience in
qualitative analysis and undergraduate laboratory instruction. You are
evaluating the reasoning trace of an autonomous agent that was asked to identify
an unknown substance through simulated experiments. Your assessment will be
entered into a benchmark report on closed-loop experimental reasoning and will
be compared across eight frontier models.

You are scoring the QUALITY OF THE REASONING PROCESS. Two things follow from
that, and they pull in opposite directions --- hold both at once:

\begin{itemize}[leftmargin=1.1em, itemsep=0.3ex, parsep=0pt, topsep=0.4ex]
\item A correct identity reached by guessing, by prior familiarity, or on
  evidence that cannot distinguish it from its neighbors is a WEAK process and
  must score low.
\item Whether the final identity is right does NOT by itself set the score. It
  gates exactly two dimensions, ESU and EGDE, through the caps in Section 6.
  CHM, DED and OIF must be scored on the reasoning alone, with no regard to
  whether the answer turned out to be correct. A model that reasoned well and
  concluded wrongly still earns those three dimensions.
\end{itemize}

Score only what the trace actually records. Do not assume the agent ``probably
did it but did not write it down''. However, an operation choice and its timing
are part of the trace and may be used as evidence of which uncertainty the agent
was addressing. A blocked operation is a normal, expected event, not an error by
itself.

EFFICIENCY --- read this before you start. Your score depends only on the final
JSON object, never on how long you deliberate. Reach the \texttt{===JSON===}
block directly and quickly: do not re-read the trace repeatedly, do not weigh
the same evidence more than once. A judgment that never emits the object is a
failed judgment, no matter how good the reasoning inside it was.

\smallskip
\noindent\textbf{SECTION 1 --- WHAT THE AGENT STARTED WITH}

\noindent Sample physical state: \texttt{\{STATE\}}\\
Initial appearance (given to the agent for free): \texttt{\{APPEARANCE\}}\\
\noindent Instruments available in this laboratory:
\begin{itemize}[leftmargin=1.1em, itemsep=0.2ex, parsep=0pt, topsep=0.3ex]
\item polarimeter (optical rotation)
\item refractometer (refractive index; liquids only)
\item conductivity meter (conductivity, mS/cm)
\item mixing bench (reagent reactions, tracked in a virtual test tube)
\item operation bench (heating, flame test, pH, magnetism, solubility)
\end{itemize}
\noindent Infrared spectroscopy and any precision melting/boiling-point
instrument are NOT available. A manual melting-range or boiling-point
observation counts as a valid but coarse measurement.

\smallskip
\noindent\textbf{SECTION 2 --- THE SESSION TRACE}

\noindent Each round lists: the agent's stated reasoning, the instruction it
issued, the safety-and-compliance gate verdict (PASS / FAIL with the reason),
the operation actually executed, and the observation returned by the bench.

A round may contain SEVERAL ``operation:'' lines, when the agent bundled
multiple steps into one instruction. Under this task prompt, combining several
independent operations in a single round is not allowed and should not be
rewarded as efficient design. Batching a step whose design depends on the result
of an earlier step in the SAME round is also a design defect.

\noindent\texttt{\{TRACE\}}

\smallskip
\noindent\textbf{SECTION 3 --- FACTS COMPUTED FROM THE RECORD (authoritative)}

\noindent These values were computed from the episode record. They are not
estimates, and you must not recompute or override them: several caps in
Section 6 are written against exactly these numbers.

\noindent Final-answer verdict: \texttt{\{VERDICT\}}

\noindent How the episode spent its rounds:

\noindent\texttt{\{TIMING\}}

\noindent Remember how the verdict may be used: it gates ESU and EGDE only.
Score CHM, DED and OIF as if you did not know it.

\smallskip
\noindent\textbf{SECTION 4 --- REFERENCE MATERIAL (never shown to the agent)}

\noindent Hidden target substance: \texttt{\{SUBSTANCE\}}\\
Its canonical, source-grounded identification route under exactly these
instruments:

\noindent\texttt{\{ROUTE\}}

\noindent Near-neighbor substances that share one or more of the target's tests
(these are the hard cases, where a non-discriminating test still returns a
plausible reading):

\noindent\texttt{\{CLUSTER\}}

\noindent Use this section to judge how much discrimination was available in
this laboratory and which test would have settled the question. It is not itself
worth points: a trace is not better because it names more of these substances,
and a test is only diagnostic if it separates the candidates the agent was
actually still considering.

\smallskip
\noindent\textbf{SECTION 5 --- PROCEDURE}

\noindent Complete Steps 1 to 3 BEFORE assigning any score in Step 4. Writing
the evidence first prevents the scores from being reverse-engineered from the
known answer.

\smallskip
\noindent\textbf{STEP 1 --- CANDIDATE AND EVIDENCE INVENTORY.} List, in order,
the candidate substances, chemical classes, or discriminating possibilities that
are visible in the trace. Visibility may come from the agent's stated reasoning,
from the operation it chose after a given observation, or from the final
exclusion reasoning. Do not require an explicit candidate list when the task
prompt forbids early candidate-set discussion.

For each visible candidate or discriminating possibility, name the observation
that introduced it, the observation that eliminated it or left it open, the
round where that happened, and --- for an elimination --- the property of that
observation that made it discriminating. Mark any asserted candidate unsupported
by observation, and any possibility that the trace keeps using after an
observation had already excluded it.

\smallskip
\noindent\textbf{STEP 2 --- EVIDENCE SUFFICIENCY OVER TIME.} Answer in two or
three sentences: at what round did the observations the agent had collected
FIRST become sufficient to support a unique or near-unique identity? Call that
round $t^*$.

Judge this on the observations available at that moment, and NOT on how the
episode turned out. In particular, do not let the declaration pull $t^*$
backwards:
\begin{itemize}[leftmargin=1.1em, itemsep=0.3ex, parsep=0pt, topsep=0.4ex]
\item if it was the LATER rounds that made the case decisive, $t^*$ is that
  later round;
\item if the agent never reached sufficient evidence, $t^*$ is ``none'' --- even
  though it declared an identity. Declaring is not evidence that the evidence
  sufficed, and a specific name announced on weak observations does not make
  those observations strong.
\end{itemize}
\noindent If a decisive discriminating test was available but never performed,
name it.

\smallskip
\noindent\textbf{STEP 3 --- DECLARATION TIMING.} State \texttt{t\_decl}, the
round in which the agent declared its final identity (or ``none''). Compare it
with $t^*$: earlier means the identity was announced before the evidence
justified it, much later means the agent kept working after it already knew
enough. Note whether any rounds between $t^*$ and \texttt{t\_decl} produced
anything useful.

\smallskip
\noindent\textbf{STEP 4 --- SCORE THE FIVE DIMENSIONS.} Every score must be an
integer, must fall on one of the anchored descriptors below, and must be
justified by a verbatim quote from the trace. When two descriptors both seem
plausible, take the LOWER one. The caps in Section 6 are hard.

\smallskip
\noindent\textbf{D1. Candidate and Hypothesis Management (CHM, 0--15)}

\noindent Did the agent manage the live possibilities in a way that did WORK ---
using observations to choose discriminating tests and to stop pursuing
possibilities that those observations had made untenable?

Judge candidate management by visible effects in the trace: stated reasoning,
operation choice, response to observations, and final exclusion logic. Do not
require explicit candidate lists, explicit narrowed-range statements, or extended
class/functionality discussion when the task prompt forbids them. A terse trace
can earn full credit if the sequence of experiments shows that observations
redirected the search. Conversely, naming many plausible substances is not
management unless those names guide or are pruned by diagnostic observations.

\begin{description}[leftmargin=1.9em, itemsep=0.5ex, parsep=0pt, topsep=0.5ex,
                    labelsep=0.5em]
\item[15] observations visibly guide a live possibility space: each major
  alternative the available tests bear on is either excluded by a discriminating
  observation, carried as still unresolved, or addressed by a later test; this
  management may be implicit in experiment choice and final exclusion logic
  rather than explicitly narrated
\item[12] observations guide the search and the live space shrinks, but some
  eliminations are only aggregate or implicit, or the pruning becomes clear only
  near the conclusion
\item[9] the agent's experiments are broadly consistent with observation-based
  possibilities, but the trace mostly accumulates evidence rather than clearly
  pruning alternatives
\item[6] the trace shows weak or loose connection between observations and the
  possibilities being tested, or a later observation should have changed the
  search but did not
\item[3] the trace mostly restates surface facts or performs generic tests, with
  little visible management of chemical possibilities
\item[0] no visible candidate or possibility management; jumps straight to a
  single identity; or continues to rely on a possibility after an observation
  had already excluded it
\end{description}

\smallskip
\noindent\textbf{D2. Diagnostic Experiment Design (DED, 0--20)}

\noindent Were the chosen experiments informative and aimed at the current
uncertainty?

\begin{description}[leftmargin=1.9em, itemsep=0.5ex, parsep=0pt, topsep=0.5ex,
                    labelsep=0.5em]
\item[20] highly diagnostic: few steps efficiently narrow the space, each aimed
  at the live candidates, nothing redundant
\item[16] targeted design working through the key properties to separate
  candidates
\item[12] most experiments eliminate something, but the ordering or the
  targeting is only fair
\item[8] a few useful experiments, but the strategy is loose and does not really
  shrink the candidate space
\item[4] mostly low-information, repeated, or with no stated discriminating
  purpose
\item[0] experiments are invalid, unexecutable, or unrelated to the
  identification
\end{description}

\noindent Notes:
\begin{itemize}[leftmargin=1.1em, itemsep=0.3ex, parsep=0pt, topsep=0.4ex]
\item a round that bundles a dependent step --- an operation whose choice could
  only be justified by the result of another operation in the SAME round, not yet
  observed when the instruction was written --- is a design defect. Do NOT apply
  this to independent measurements batched together (e.g.\ pH and conductivity on
  the same sample), which is legitimate practice.
\item a non-discriminating test repeated after it already returned its answer
  counts against this dimension.
\item reducing the step count is itself part of good design here: reaching the
  same discrimination in fewer rounds is better, not merely equivalent.
\end{itemize}

\smallskip
\noindent\textbf{D3. Observation Interpretation Fidelity (OIF, 0--15)}

\noindent Did the agent interpret the bench's returns faithfully and chemically
correctly?

\begin{description}[leftmargin=1.9em, itemsep=0.5ex, parsep=0pt, topsep=0.5ex,
                    labelsep=0.5em]
\item[15] every key interpretation is accurate, restrained and traceable;
  negative results and residual uncertainty are handled correctly
\item[12] faithful and chemically sound; key conclusions trace back to
  observations
\item[9] mostly consistent with the observations, with some mild over-reach or
  loose wording
\item[6] some interpretations correct, but several unsupported claims or a clear
  chemical error
\item[3] frequent reference to phenomena that were not observed, wrong
  attributions, or serious over-interpretation
\item[0] most interpretations contradict the observations, or the core reasoning
  rests on fabricated phenomena
\end{description}

\noindent Caps:
\begin{itemize}[leftmargin=1.1em, itemsep=0.3ex, parsep=0pt, topsep=0.4ex]
\item the agent repeatedly cites observations the bench never returned
  $\rightarrow$ OIF at most 6
\item a ``no visible change'' or ``cannot determine'' return is used as a
  confirmed positive or a confirmed exclusion $\rightarrow$ OIF at most 9
\item a qualitative result is upgraded to a definite structure or a definite
  number with nothing in the trace to support it $\rightarrow$ OIF at most 9
\end{itemize}

\smallskip
\noindent\textbf{D4. Evidential Sufficiency and Uniqueness (ESU, 0--30)}

\noindent On the observations this episode actually produced, is the declared
identity uniquely supported, and are the main competing candidates ruled out?
THIS IS NOT ACCURACY. A correct answer on non-unique evidence scores LOW here.

\begin{description}[leftmargin=1.9em, itemsep=0.5ex, parsep=0pt, topsep=0.5ex,
                    labelsep=0.5em]
\item[30] the accumulated observations uniquely support the declared substance
  and explicitly rule out the main competing candidates
\item[24] the evidence essentially supports the declared substance; only minor
  gaps or slight residual uncertainty
\item[18] the evidence supports it fairly strongly, but an important competing
  candidate was never excluded
\item[12] the evidence supports the declared substance, but equally supports
  several reasonable candidates that remain unresolved
\item[6] the evidence supports only a very broad class (``a carboxylic acid'',
  ``an aldehyde'', ``a sodium salt''), not a specific substance
\item[0] the conclusion contradicts the key observations, or rests on no
  evidence
\end{description}

\noindent Judge the evidence itself, not the agent's confidence in it, and not
whether the declared name happens to be right. Reliance on prior familiarity or
on which substance is common, rather than on this episode's observations, is a
reason to stay at a low level. The caps in Section 6 are binding on this
dimension.

\smallskip
\noindent\textbf{D5. Evidence-Gated Decision Efficiency (EGDE, 0--20)}

\noindent Once the evidence was sufficient, did the agent conclude --- or did it
keep going?

This is NOT a speed reward. A fast guess is worth nothing: a short episode spent
on a guess is not better than a long one spent gathering evidence. The whole
dimension is the comparison between the two rounds you established in Steps 2
and 3 --- $t^*$, when the evidence first sufficed, and \texttt{t\_decl}, when the
agent said what the substance was. In this benchmark an episode ends when the
identity is stated, so \texttt{t\_decl} is always the last round; what varies is
how much work preceded it that $t^*$ had already made unnecessary.

\begin{description}[leftmargin=1.9em, itemsep=0.5ex, parsep=0pt, topsep=0.5ex,
                    labelsep=0.5em]
\item[20] concluded one round after $t^*$, with nothing wasted after $t^*$
\item[16] concluded two rounds after $t^*$
\item[12] concluded three or more rounds after $t^*$
\item[8] concluded three or more rounds after $t^*$, and most of those extra
  rounds returned no new observation; a wrong final answer normally sits at this
  level or below
\item[4] $t^*$ never existed --- a specific identity was named on evidence that
  never became sufficient to support it
\item[0] no valid final declaration; or the conclusion is a pure guess with no
  evidence; or it plainly contradicts the observations
\end{description}

\noindent Think of it as: the agent had \texttt{\{BUDGET\}} rounds. The moment it
had enough to name the substance, every further round is a round it did not need.
Credit here is for recognizing sufficiency, and it is only available to the
extent D4 says the evidence really was sufficient.

\smallskip
\noindent\textbf{SECTION 6 --- GLOBAL CAPS}

\noindent Apply these on top of the per-dimension caps; the tightest cap wins.
Where a cap names a condition you can read in the record, take the verdict and
the timing in Section 3 as settled fact. Where it names a judgment, make it
yourself.

\smallskip
\noindent\textbf{ESU}
\begin{enumerate}[leftmargin=1.6em, itemsep=0.3ex, parsep=0pt, topsep=0.4ex]
\item No valid final declaration $\rightarrow$ ESU at most 6, and usually 0.
\item Declared identity is wrong (Section 3 verdict) $\rightarrow$ ESU at most
  12. If the conclusion also contradicts a key observation $\rightarrow$ ESU
  must be 0 or 6.
\item The evidence supports only a broad class rather than a specific substance
  $\rightarrow$ ESU at most 12, even when the declared name is correct.
\item A major competing candidate remained unexcluded $\rightarrow$ ESU at most
  18.
\item The agent repeatedly invokes observations the bench never returned
  $\rightarrow$ ESU at most 12, and OIF at most 6.
\end{enumerate}

\smallskip
\noindent\textbf{EGDE}
\begin{enumerate}[leftmargin=1.6em, itemsep=0.3ex, parsep=0pt, topsep=0.4ex]
\item No valid final declaration $\rightarrow$ EGDE must be 0.
\item Declared identity is wrong (Section 3 verdict) $\rightarrow$ EGDE at most
  8. If the conclusion also contradicts a key observation $\rightarrow$ EGDE
  must be 0 or 4.
\item EGDE may never exceed what the ESU you just awarded can pay for:
  \begin{center}
  \begin{tabular}{@{}ll@{}}
    ESU 0 or 6 $\rightarrow$   & EGDE at most 4 \\
    ESU 12 $\rightarrow$       & EGDE at most 8 \\
    ESU 18 $\rightarrow$       & EGDE at most 12 \\
    ESU 24 or 30 $\rightarrow$ & EGDE may be 16 or 20 \\
  \end{tabular}
  \end{center}
  A model does not get decision-efficiency credit for concluding quickly on
  evidence that does not support the conclusion.
\item The declaration came only in the final round (\texttt{\{BUDGET\}})
  $\rightarrow$ EGDE at most 16, even if the answer is correct and the evidence
  sufficient; at most 8 if the evidence was not sufficient.
\item Note that EGDE carries no cap for the share of rejected or repeated steps.
  That share is a separate matter with its own place in the benchmark, and it
  does not enter this dimension: credit here is for concluding when the evidence
  allowed it, not for the gate's opinion of the work. The timing bands --- how
  many rounds late or early the declaration came --- are the descriptors of D5;
  apply them there rather than restating them here as a cap.
\end{enumerate}

\noindent List every cap you actually applied in ``caps\_applied'', using these
names: \texttt{NO\_FINAL\_DECLARATION}, \texttt{INCORRECT\_ANSWER},
\texttt{CONTRADICTS\_OBSERVATION}, \texttt{CLASS\_ONLY},
\texttt{COMPETITORS\_UNEXCLUDED}, \texttt{HALLUCINATION},
\texttt{DECLARED\_AT\_BUDGET}, \texttt{ESU\_GATED\_EGDE}. If none applied, use
\texttt{["NONE"]}. Do not list a cap you did not apply, and do not apply one you
did not list.

\smallskip
\noindent\textbf{SECTION 7 --- HOW TO REPLY}

\noindent Reply in exactly TWO parts.

\noindent\textbf{PART 1 --- ANALYSIS.} Plain prose, at most about 350 words. Work through Steps 1 to 4 above: the visible candidate/evidence or
possibility/evidence inventory, the evidence-sufficiency answer with $t^*$, the
declaration timing, then each of the five scores with the trace quote or
operation choice you are basing it on. This part is read by
human reviewers and is part of the record. Be concise --- a long deliberation
that never reaches the JSON is a failed judgment.

\noindent\textbf{PART 2 --- MACHINE-READABLE RESULT.} After the analysis, write
the line

\begin{center}
\texttt{===JSON===}
\end{center}

\noindent on a line by itself, then the JSON object below, and NOTHING after it.
The JSON object must be the last thing in your reply. No markdown fences, no
comments inside the JSON. Every string must be valid JSON; keep each
justification under 200 characters.

{\footnotesize
\begin{verbatim}
{
  "scores": {
    "candidate_and_hypothesis_management":
      <int 0-15>,
    "diagnostic_experiment_design":
      <int 0-20>,
    "observation_interpretation_fidelity":
      <int 0-15>,
    "evidential_sufficiency_uniqueness":
      <int 0-30>,
    "evidence_gated_decision_efficiency":
      <int 0-20>
  },
  "tds": <int 0-100, the sum of the
          five scores>,
  "justifications": {
    "candidate_and_hypothesis_management":
      "<one sentence>",
    "diagnostic_experiment_design":
      "<one sentence>",
    "observation_interpretation_fidelity":
      "<one sentence>",
    "evidential_sufficiency_uniqueness":
      "<one sentence>",
    "evidence_gated_decision_efficiency":
      "<one sentence>"
  },
  "t_star": <int round 1-10 at which the
             evidence first became
             sufficient, or null>,
  "t_decl": <int round 1-10 in which the
             identity was declared, or
             null>,
  "caps_applied": ["<NONE, or one or
                    more of the cap names
                    listed in Section 6>"],
  "one_line_verdict": "<one sentence,
                       under 30 words>"
}
\end{verbatim}}
}
\end{prompt}

\end{document}